\documentclass[lettersize,journal]{IEEEtran}
\usepackage{amsmath,amsfonts}
\usepackage{algorithmic}
\usepackage{array}
\usepackage[caption=false,font=normalsize,
labelfont=sf,textfont=sf]{subfig}
\usepackage{textcomp}
\usepackage{stfloats}
\usepackage{url}
\usepackage{verbatim}
\usepackage{graphicx}
\usepackage{balance}

\usepackage{algorithm}
\usepackage{algorithmic}
\usepackage{cite}
\usepackage[breaklinks]{hyperref}
\usepackage{booktabs}
 
\begin{document}
\title{Multimodal Dreaming: A Global Workspace Approach to \\ World Model-Based Reinforcement Learning}

\author{L\'eopold~Mayti\'e \& Roland~Bertin~Johannet \& Rufin~VanRullen
\IEEEcompsocitemizethanks{\IEEEcompsocthanksitem Univ Toulouse, CNRS, CerCo, and ANITI, Artificial and Natural Intelligence Toulouse Institute
\IEEEcompsocthanksitem Correspondence: rufin.vanrullen@cnrs.fr}
}

\maketitle

\begin{abstract}
Humans leverage rich internal models of the world to reason about the future, imagine counterfactuals, and adapt flexibly to new situations. In Reinforcement Learning (RL), world models aim to capture how the environment evolves in response to the agent's actions, facilitating planning and generalization. However, typical world models directly operate on the environment variables (e.g. pixels, physical attributes), which can make their training slow and cumbersome; instead, it may be advantageous to rely on high-level latent dimensions that capture relevant multimodal variables. Global Workspace (GW) Theory offers a cognitive framework for multimodal integration and information broadcasting in the brain, and recent studies have begun to introduce efficient deep learning implementations of GW. Here, we evaluate the capabilities of an RL system combining GW with a world model. We compare our GW-Dreamer with various versions of the standard Proximal Policy Optimization (PPO) and the original Dreamer algorithms. We show that performing the dreaming process (i.e., mental simulation) inside the GW latent space allows for training with fewer environment steps. As an additional emergent property, the resulting model (but not its comparison baselines) displays strong robustness to the absence of one of its observation modalities (images or simulation attributes). We conclude that the combination of GW with World Models holds great potential for improving decision-making in RL agents.
\end{abstract}


\section{Introduction}

\IEEEPARstart{H}umans possess the ability to anticipate the consequences of their actions before executing them in the real world. This capacity suggests that humans construct an internal World Model (see e.g. \cite{friston_free-energy_2010,clark_whatever_2013}, among others).

In artificial intelligence (AI), this concept has been particularly applied in World Model-based reinforcement learning (RL), a subset of model-based RL. In model-based RL, transition dynamics of the environment are traditionally specified as Markov decision processes (MDPs), either manually defined \cite{sutton_dyna_1991, atkeson_comparison_1997} or empirically estimated from interaction data \cite{dearden_bayesian_nodate, szita_model-based_nodate}. While model-based RL is generally more sample-efficient than model-free RL, constructing accurate transition models remains challenging. Learning a World Model directly from data facilitates decision-making in environments where transition dynamics are either unknown or too complex to specify explicitly~\cite{ha_recurrent_2018}. The recently published DreamerV3 framework \cite{hafner_mastering_2023}, for instance, showcases faster convergence and improved performance, by allowing agents to learn within mental simulations of episodes rather than relying solely on direct interaction with the environment.
More generally, AI research on world models has gained further momentum with the advent of large-scale World Foundation Models, such as Genie \cite{bruce_genie_2024} trained in an unsupervised manner, or World Foundation Models platforms like Cosmos from Nvidia \cite{nvidia_cosmos_2025}

While World Models can improve efficiency in solving complex tasks, they are typically trained with high-dimensional data reconstruction objectives ~\cite{ha_recurrent_2018, hafner_mastering_2023}. However, reconstructing high-dimensional data such as images has been criticized due to computational costs and reconstruction complexity. Consequently, alternative approaches have emerged using compact, low-dimensional representations~\cite{garrido_learning_2024, bardes_revisiting_2024, assran_v-jepa_2025}.
Reconstructing high-dimensional data in multimodal environments (e.g. vision, audio, language) becomes even more computationally expensive and complex. This contrasts with humans’ abilities to perceive the world through multiple sensory modalities, to create a rich and robust representation of their environment and to anticipate the consequences of their actions in this multimodal context. This contrast motivates taking inspiration from human cognition.

The Global Workspace Theory (GWT), introduced by~\cite{baars_cognitive_1988} and later expanded by~\cite{dehaene_neuronal_1998}, provides a framework to explain multimodal integrative cognitive processes. According to this theory, specialized modules compete to encode their information into a shared space called the Global Workspace. Through a broadcasting mechanism, this information becomes accessible to various brain regions, shaping our conscious experience.
Deep Learning implementations of this theory have demonstrated strong multimodal grounding capabilities for downstream tasks~\cite{devillers_semi-supervised_2024, maytie_zero-shot_2024}; this suggests that such multimodal low-dimensional representations could be used advantageously in model-based RL systems like Dreamer.

Therefore, in this paper we introduce a system bridging the ideas of World Model and Global Workspace \cite{vanrullen_deep_2021, safron_integrated_2022}. We took inspiration from the architecture proposed in Dreamer algorithms \cite{ha_recurrent_2018, hafner_mastering_2023} and extended the Global Workspace implementation proposed by \cite{devillers_semi-supervised_2024} to implement our Global Workspace Dreamer (GW-Dreamer). The key originality of GW-Dreamer is that it learns to represent the World-Model transitions using multimodal GW representations. We compared our model in two different Reinforcement Learning environments against standard Proximal Policy Optimization (PPO) and the original DreamerV3 algorithm~\cite{hafner_mastering_2023} as well as two variants of Dreamer, one that shared the same visual input module as GW-Dreamer, and another that replaced the multimodal GW representation with a CLIP-like representation\cite{radford_learning_2021}. Thanks to its efficient GW multimodal latent representation, our model learns with fewer environment interactions; in addition, it proves more robust to missing modalities (as already shown in the context of model-free RL by \cite{maytie_zero-shot_2024}).

\section{Model}
\label{sec:Model}

In this study, we consider RL environments with multimodal observations. By consequence, the state of an environment at time t leads to multiple observations $o_t \in \mathcal{O}$, which can be either an RGB image $o_t^v$, or an attribute vector $o_t^{attr}$ describing physical attributes of the simulation. From these observations, the agent predicts an action $a_t \in \mathcal{A}$ to interact with the environment, leading to a reward $r_{t+1}$.

To interact with these multimodal environments we propose a model composed of three main components: a representation model, called Global Workspace, a World Model and an Actor-Critic RL policy. The training process consists of two main steps. First, the Global Workspace is trained to represent the multimodal environment using a dataset of environment observations collected randomly or via an expert agent (Figures~\ref{Fused_GW} and~\ref{Fused_GW_alphas}). Then, the World Model and the Actor-Critic are trained through interaction with the environment (Figure \ref{GW_Dreamer}). In the following subparts, we provide a description of the architecture and training procedure for each component; details and hyperparameters can be found in the Appendix, Tables \ref{table:VAE_v_SS} to \ref{table:Dreamer_training}

\subsection{Global Workspace}

\begin{figure*}[htb!]
    \begin{center}
        \includegraphics[scale=0.4]{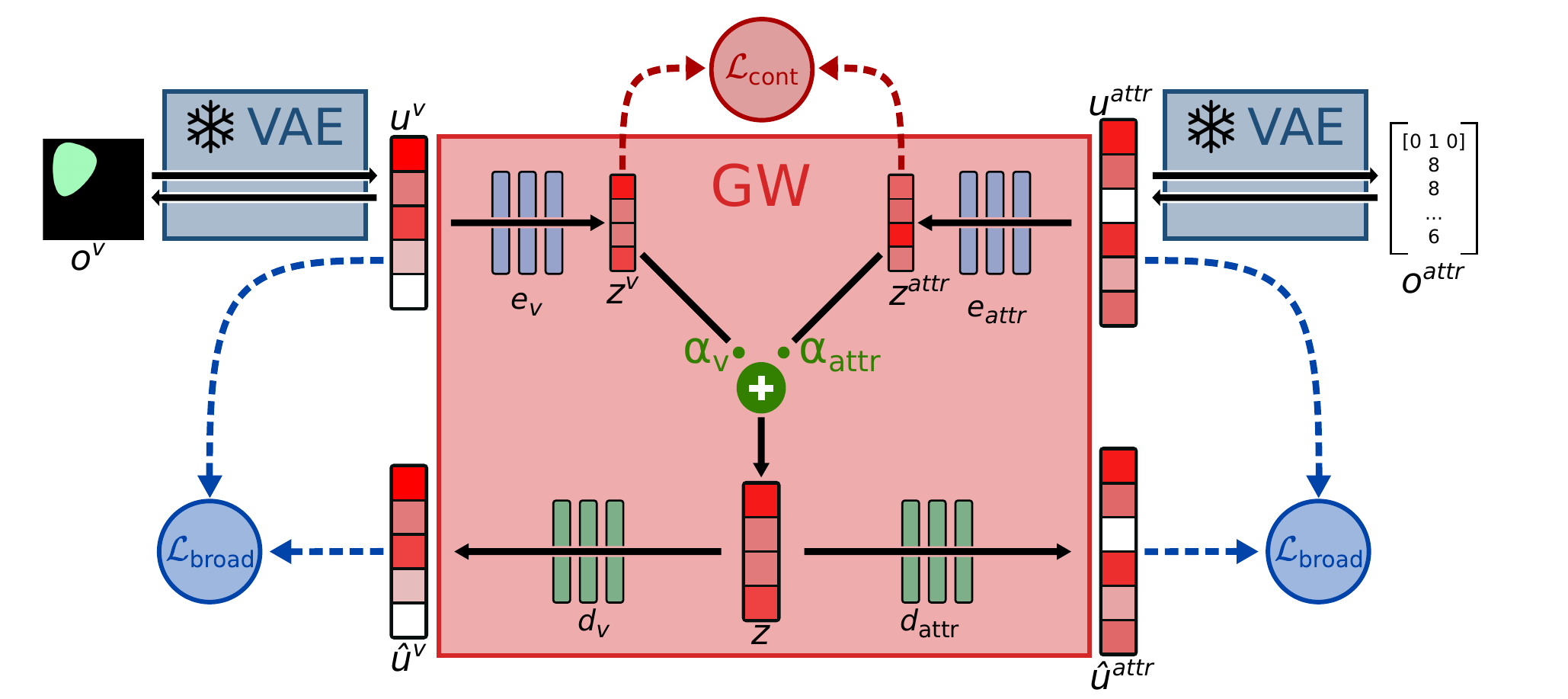}
    \end{center}
    \caption{Overview of the Global Workspace model for multimodal representation. Raw environment inputs (image pixels, simulation attributes) are encoded in their latent unimodal representation ($u^{v}$ or $u^{attr}$) thanks to pretrained (and frozen) VAEs. These unimodal latent representations are then processed by encoders $e_v$ and $e_{attr}$ (respectively) to produce pre-GW representations ($z^v$ and $z^{attr}$). The final Global Workspace representation $z \in \mathcal{Z}$ is obtained by fusing these pre-GW representations through an element-wise weighted sum (with weights $\alpha_v\geq 0$ and $\alpha_{attr}\geq 0$, $\alpha_v+\alpha_{attr}=1$) followed by a Tanh activation. The unimodal latent vectors can be retrieved from $z$ with a set of decoders $d_v$ and $d_{attr}$. The GW component networks $e_{v}$, $e_{attr}$, $d_{v}$ and $d_{attr}$ are trained by combining a contrastive loss $\mathcal{L}_{cont}$ and a broadcast $\mathcal{L}_{broad}$. The former encourages the pre-GW representations to align across modalities; the latter also promotes this objective (see \cite{devillers_semi-supervised_2024}), and ensures that decoded or ``broadcasted'' GW representations resemble the original unimodal latent representations, regardless of each modality's initial contribution to the GW representation (as captured by the fusion weights $\alpha_v$ and $\alpha_{attr}$). The GW module can be trained jointly with the rest of the model or pre-trained and subsequently frozen during the learning of the World Model and RL policy (Figure~\ref{GW_Dreamer}), using fixed fusion weights ($\alpha_v = \alpha_{attr} = 0.5$).} 
    \label{Fused_GW}
\end{figure*}

\begin{figure*}[htb!]
    \begin{center}
        \includegraphics[scale=0.36]{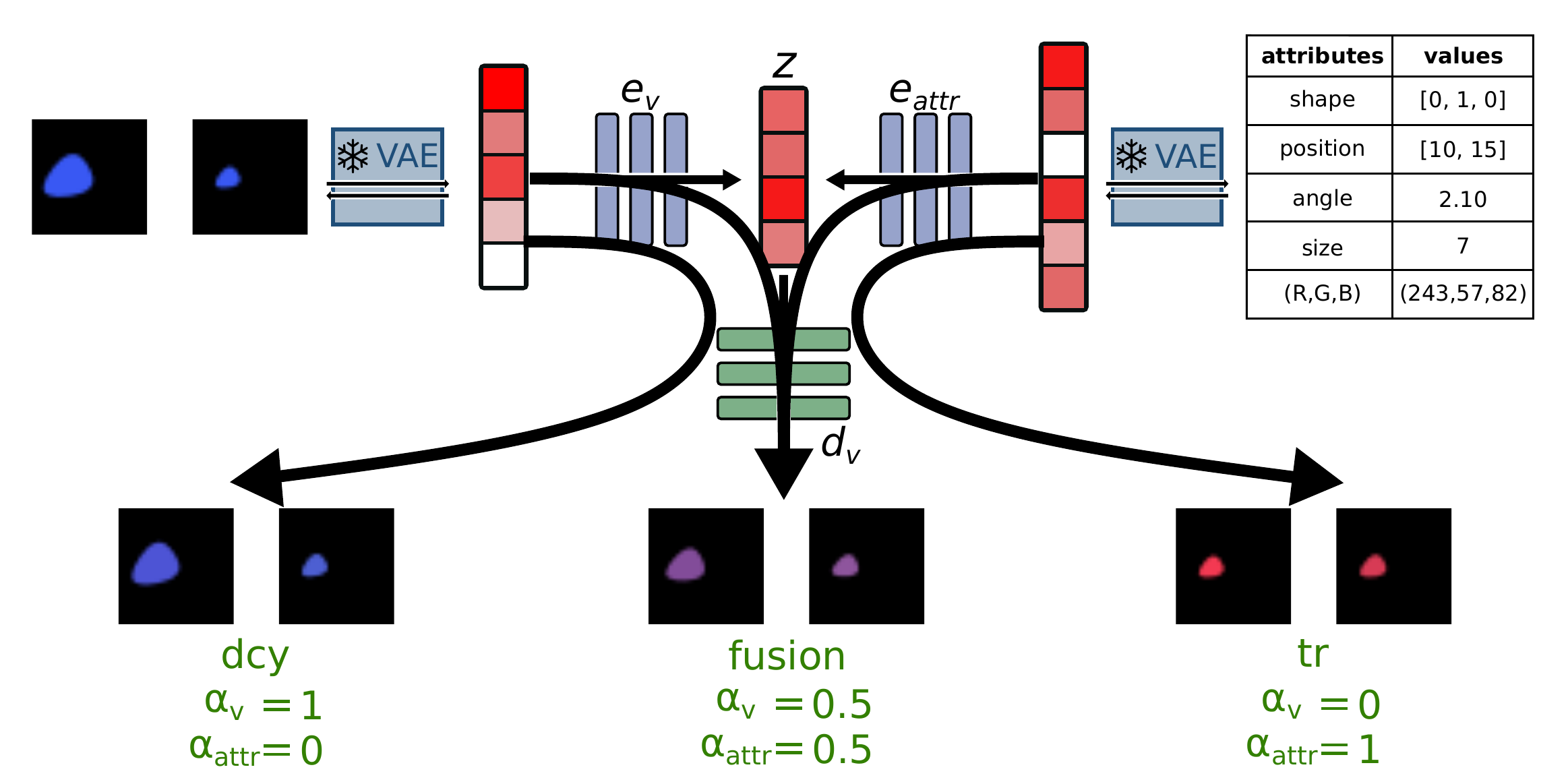}
    \end{center}
    \caption{Illustration of the behaviour of the GW. We start from a fixed attribute vector describing a small red egg-shape (top right), and two images (top left) that are chosen to be incongruent with the attribute vector, in terms of color (right-most image) or both color and size (left-most image). These inputs are encoded into the GW using different fusion weights $\alpha_v$ and $\alpha_{attr}$, indicated in green below each configuration, and subsequently decoded into an image. The resulting images at the bottom illustrate three distinct functional modes of operation. \textbf{In the translation mode (tr, bottom right)}, both modalities are encoded, but only attribute information is transmitted through the GW, while visual input is disregarded. The reconstructed images, obtained by decoding the GW latent vector $z$ as an image using $d_v$ and the visual VAE, demonstrate the successful translation of attribute information into the visual domain (both objects are small and red). \textbf{In the demi-cycle mode (dcy, bottom left)}, both modalities are encoded, but only the visual information is propagated through the GW. The absence of distortions due to attributes information in the reconstructed images confirms that attribute information was effectively suppressed. \textbf{In the fusion mode (bottom middle)}, both modalities are encoded with equal weights, allowing information from both sources to be integrated inside the GW. The decoded images reflect a hybrid representation of vision and attributes features, resulting in an intermediate color and size. } 
    \label{Fused_GW_alphas}
\end{figure*}

The Global Workspace serves as a representation model, encoding multiple modalities into a unified latent representation. This latent representation is then used by the World Model to encapsulate the agent’s perception of the environment at time t. 
Our proposed Global Workspace is inspired by the approach introduced by \cite{devillers_semi-supervised_2024}. However, we modify both the architecture and training procedure to produce a single unified representation, rather than maintaining separate representations for each modality. This architecture and its training losses are illustrated in Figure \ref{Fused_GW}.

As proposed by \cite{vanrullen_deep_2021} and implemented by \cite{devillers_semi-supervised_2024}, we do not train our set of encoders and decoders directly from raw modalities. Instead, we employ two pretrained and frozen modules (in this case, VAEs) to transform raw representations (denoted $o^v$ for images and $o^{attr}$ for attributes) into unimodal latent representations ($u^v$ and $u^{attr}$). These unimodal representations are then encoded into pre-fusion latent variables $z^v=e_v(u^v)$ and $z^{attr}=e_{attr}(u^{attr})$. However, in contrast to the model proposed by Devillers~\cite{devillers_semi-supervised_2024}, we do not always directly decode from these pre-fusion representations. We combine them using element-wise weighted sum followed by a Tanh activation function to form a unified representation denoted $z$. From this unified representation, we can recover the unimodal representations through a set of decoders $d_v$ and $d_{attr}$.

The GW model is trained using two loss functions: the contrastive loss and the broadcast loss. The contrastive loss $\mathcal{L}_{cont}$ follows a similar formulation to the one proposed in CLIP \cite{radford_learning_2021}; it is designed to align the representations $z^v$ and $z^{attr}$ before fusion, supporting the development of amodal representations in the Global Workspace. The broadcast loss is inspired by the broadcast principle at the heart of GWT; it is computed by comparing the predicted ($\hat{u}^v$, $\hat{u}^{attr}$) and ground-truth unimodal representations ($u^v$, $u^{attr}$) using the mean squared error (MSE). Specifically, it consists of a weighted sum of multiple sub-losses, including the cycle loss ($\mathcal{L}_{cy}$), demi-cycle loss ($\mathcal{L}_{dcy}$), and translation loss ($\mathcal{L}_{tr}$), as introduced by \cite{devillers_semi-supervised_2024} and described in equations \ref{equation:gw_losses} and \ref{L_broadcast}. Additionally, a fusion loss ($\mathcal{L}_{fusion}$) is incorporated. The primary objective of these losses is to ensure that the unimodal latent vectors can be accurately reconstructed after the weighted-sum fusion of pre-Global Workspace representations ($z^v$ and $z^{attr}$), regardless of the exact fusion weights employed. Figure ~\ref{Fused_GW_alphas} illustrates the effect of changing these weighted-sum coefficients, thus allowing more or less information into the workspace from one modality relative to the other. The weighting factors $\alpha_{attr}$ and $\alpha_v$ are adjusted to compute each specific sub-loss (and with the constraints $\alpha_v \geq 0, \alpha_{attr} \geq 0,  \alpha_v + \alpha_{attr} = 1$), as defined in equation \ref{equation:gw_losses}. (For an in-depth discussion of the usefulness of each loss term $\mathcal{L}_{cont}$, $\mathcal{L}_{cy}$, $\mathcal{L}_{dcy}$ and $\mathcal{L}_{tr}$, see \cite{devillers_semi-supervised_2024}).

\begin{equation}
    \label{equation:gw_losses}
    \begin{array}{c}
        \forall (x,y) \in \{attr,v\}, \quad x \neq y \\[8pt]
        \left\{
            \begin{aligned}
              & \mathcal{L}_{dcy} = \|d_x(tanh(e_x(u^x))) - u^x \|^2_2, \\
              & \hspace{1 cm} \scriptstyle{(\alpha_x = 1, \alpha_{y} = 0)}\\
              & \mathcal{L}_{cy} = \|d_x(tanh(e_y(d_y(tanh(e_x(u^x)))))) - u^x\|^2_2, \\
              & \hspace{1 cm} \scriptstyle{(\alpha_x = 1, \alpha_{y} = 0), \text{ then } (\alpha_x = 0, \alpha_{y} = 1)}\\
              & \mathcal{L}_{tr} = \|d_y(tanh(e_x(u^x))) - u^y\|^2_2,\\
              & \hspace{1 cm} \scriptstyle{(\alpha_x = 1, \alpha_{y} = 0)}\\
              & \mathcal{L}_{fusion} = \|d_x(tanh(\alpha_x .e_x(u^x) + \alpha_y .e_y(u^y))) - u^x\|^2_2, \\
              & \hspace{1 cm} \scriptstyle{(\alpha_x > 0, \alpha_{y} > 0, \alpha_x + \alpha_{y} = 1)}\\
            \end{aligned}
        \right.\\
    \end{array}
\end{equation}

\begin{equation}
    \label{L_broadcast}
    \begin{aligned}
      & \mathcal{L}_{broad} = \beta_{dcy} . \mathcal{L}_{dcy} + \beta_{cy} . \mathcal{L}_{cy} + \beta_{tr} . \mathcal{L}_{tr} + \beta_{fusion} . \mathcal{L}_{fusion} \\
    \end{aligned}
\end{equation}

The full training objective of the Global Workspace is a weighted sum of the contrastive loss and the broadcast loss, as shown in equation \ref{L_GW}. The weight of the broadcast loss is fixed at one (its overall contribution can be adjusted by modifying the individual weights that constitute it: $\beta_{dcy}, \beta_{cy}, \beta_{tr}, \beta_{fusion}$).

\begin{equation}
    \label{L_GW}
    \begin{aligned}
      & \mathcal{L}_{GW} = \beta_{cont} . \mathcal{L}_{cont} + \mathcal{L}_{broad} \\
    \end{aligned}
\end{equation}

Using different combinations of losses can lead to model variants with different properties in terms of self-supervision, Global Workspace broadcast or multimodal alignment. One variant is particularly interesting as a comparison baseline: keeping only the contrastive loss during training by setting all other loss weights to zero ($\beta_{dcy} = \beta_{cy} = \beta_{tr} = \beta_{fusion} = 0$) leads to a model close to CLIP~\cite{radford_learning_2021} that can be named ``CLIP-like''. This baseline retains the exact same architecture as our Global Workspace (GW) model but is optimized with only the contrastive component, enabling a direct comparison by isolating the effect of our full training objective while maintaining architectural consistency.

Figure \ref{Fused_GW_alphas} illustrates the functional properties of the trained GW when processing multimodal inputs. Here, for illustration purposes, two images (left) are chosen to differ from the attributes vector (right) along the color and size dimensions. By modulating the fusion coefficients $\alpha_v$ and $\alpha_{attr}$, the GW can perform distinct functional operations. For instance, it can perform attribute-to-image \textit{translation} by setting the fusion coefficient to $\alpha_v = 0$ and $\alpha_{attr} = 1$. This configuration ensures that only the attributes representation propagates through the GW, while visual information is disregarded. The following decoding step (using $d_v$ and the visual VAE), reconstructs an image that is inferred purely from the attribute representation. The results, shown above the "tr" label in Figure \ref{Fused_GW_alphas}, confirm that the reconstructed image remains unaffected by the suppressed visual input.
A second operational mode, referred to as a \textit{demi-cycle}, is illustrated in the “dcy” section of Figure \ref{Fused_GW_alphas}. Here, the GW selectively propagates only the visual representation by setting $\alpha_v = 1$ and $\alpha_{attr} = 0$. The reconstructed images exclusively reflect the visual inputs.
Beyond unimodal processing, the GW also supports multimodal \textit{fusion}, enabling the integration of heterogeneous information streams. In the "fusion" section of Figure \ref{Fused_GW_alphas}, both modalities are encoded into the GW with equal weighting ($\alpha_v = 0.5,\alpha_{attr} = 0.5$), allowing information from both sources to be jointly encoded in the shared latent space. Decoding this fused representation results in an image that integrates features from both the original visual input and attributes: the reconstructed color and size are halfway between those of the attributes and image inputs. Thus, by appropriately tuning the fusion coefficients following encoding ($e_v, e_{attr}$) and by selecting the appropriate decoding pathway ($d_v$ or $d_{attr}$), one can dynamically reconfigure the functional role of the GW. Specifically, it can transition between unimodal reconstruction, cross-modal translation, and multimodal fusion, providing a flexible framework for integrating and transforming heterogeneous information sources.

\subsection{World Model}

The World Model (WM) is a fundamental component that enables training the RL agent through a mental simulation or ``dreaming'' process. At each time step $t$, this recurrent network (with internal hidden state $h_t$) receives as input the GW representation ($z_t$) and an action $a_t$, from which it predicts the GW representation at the next time step ($\hat{z}_{t+1}$), the reward associated with the action ($\hat{r}_{t+1}$), and a Boolean termination signal ($\hat{d}_{t+1}$) indicating whether the task is complete. (The hat notation is used to indicate that the variable represents a predicted next observation from the world model, rather than an observation from the environment).The GW representation is computed from environmental observations using the pre-trained (and frozen) GW model described above, with constant fusion weights of 0.5 ($\alpha_{attr} = \alpha_v = 0.5$). The model employs a Gated Recurrent Unit (GRU) \cite{cho_learning_2014} for recurrence, while the prediction heads consist of a set of Multi-Layer Perceptrons (MLPs).

The World Model is trained on data collected from the environment using the current policy of the Actor (see below), while keeping the GW model frozen. Its objective is to predict the GW representation, reward, and termination flag at the next time step (t+1). The corresponding loss function ($\mathcal{L}_{WM}$) is computed as a weighted sum, combining MSE for the predicted GW representation and reward, with Binary Cross-Entropy (BCE) for the termination flag. This is illustrated in Figure \ref{GW_Dreamer} (1).

\subsection{Actor-Critic}

\begin{figure*}[tb!]
    \begin{center}
        \includegraphics[scale=0.35]{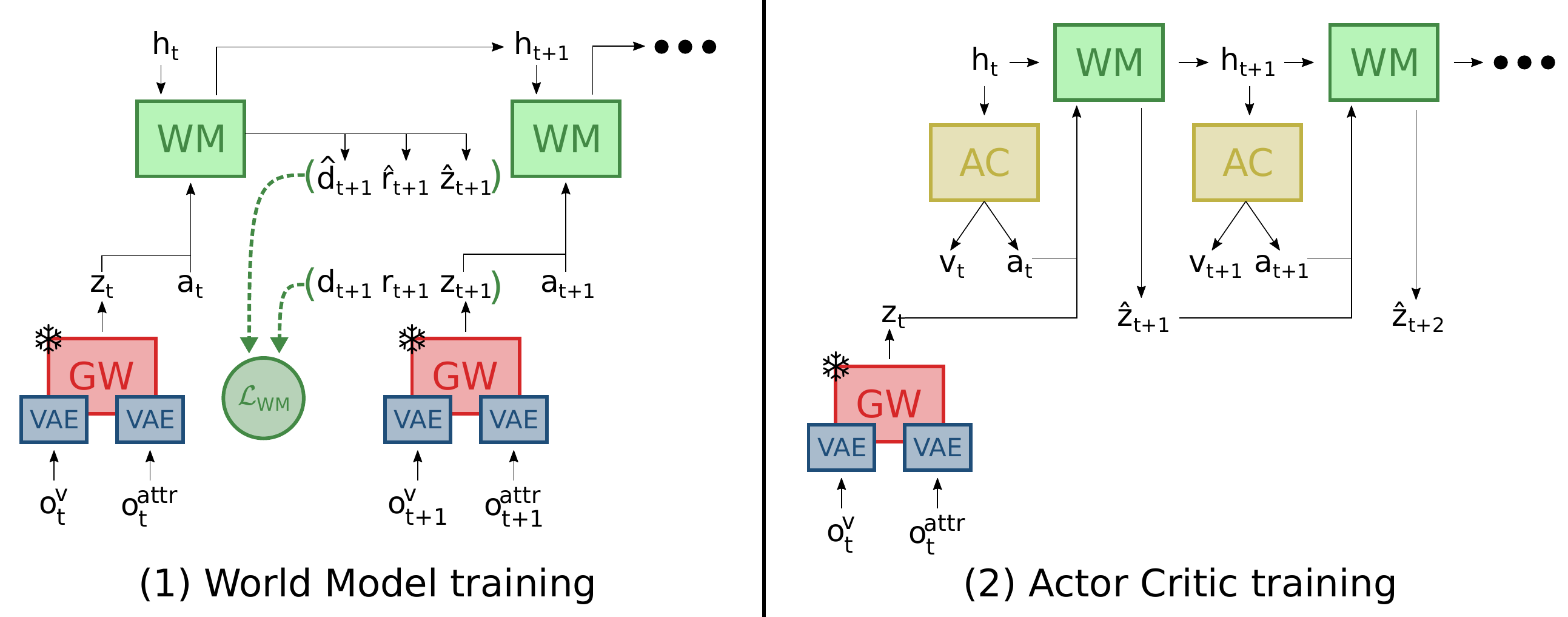}
    \end{center}
    \caption{\textbf{(1) World Model training}: At each time step, the environment provides observations ($o^v_t$, $o^{attr}_t$), a reward $r_t$, and a termination signal $d_t$. A pretrained and frozen Global Workspace (GW) model, incorporating a Variational Autoencoder (VAE) for each modality, encodes observations into a GW representation $z_t$. The WM is trained on sequences of data collected from the environment using the current AC policy. Given $z_t$ and the action $a_t$ predicted by the policy, the WM (implemented as a GRU: Gated Recurrent Unit) updates its internal state from $h_t$ to $h_{t+1}$. Using this updated state, the WM predicts the next GW representation $z_{t+1}$, the expected reward $r_{t+1}$, and the termination signal $d_{t+1}$ with three separate prediction heads. The loss function $\mathcal{L}_{WM}$ is computed as a weighted sum of the Mean Squared Error (MSE) for $z_{t+1}$ and $r_{t+1}$, and the Binary Cross-Entropy (BCE) loss for predicting $d_{t+1}$. \textbf{(2) Actor-Critic training}: The AC model is trained using "mental simulation". The GW representation $z_t$ derived from observations is provided only at the first time step. For subsequent steps, the WM generates novel states by processing the previously predicted GW representation and the action selected by the AC. The AC loss functions are computed exclusively from the predicted elements within the simulated trajectory, including the generated termination signal $\hat{d}$, reward $\hat{r}$, and actions taken based on the latent state $h$.} 
    \label{GW_Dreamer}
\end{figure*}

The Actor policy is learned concurrently with the World Model in alternating steps, as detailed below.
Initially, $n$ data pairs ($o_t$, $a_t$, $o_{t+1}$, $r_{t+1}$, $d_{t+1}$) are collected through interaction with the environment and stored in a replay buffer. Once the data is gathered, $m$ learning steps are performed. The learning alternates between training the World Model as described previously and the Actor-Critic network. The Actor-Critic takes as input the internal representation from the GRU, $h_t$, and predicts both the action $a_t$ and the state value $v_t$. This approach closely follows the methodology proposed in Dreamer \cite{hafner_mastering_2023}. It learns from "mental simulations" or "dreaming" instead of interacting directly with the environment. During training, the observations are provided only at the first step. For the following ones, the World Model simulates the environment for a predetermined number of steps, using the actions predicted by the Actor, without access to true observations, as shown in Figure \ref{GW_Dreamer} (2). During AC training, the gradient does not propagate through the World Model, ensuring that the learned policy does not directly influence the internal dynamics of the World Model (and vice-versa, WM training gradients do not propagate to the AC network). The simulated actions and values, along with the predicted rewards and done signals, are used in the loss function of the Actor-Critic network (following the standard Actor-Critic algorithm \cite{konda_actor-critic_1999} modified in Dreamer \cite{hafner_mastering_2023}). This entire training procedure is described explicitly in Algorithm \ref{algo:training_algorithm} in Appendix.

\section{Environments}

\begin{figure*}[htb!]
    \begin{center}
        \includegraphics[scale=0.2]{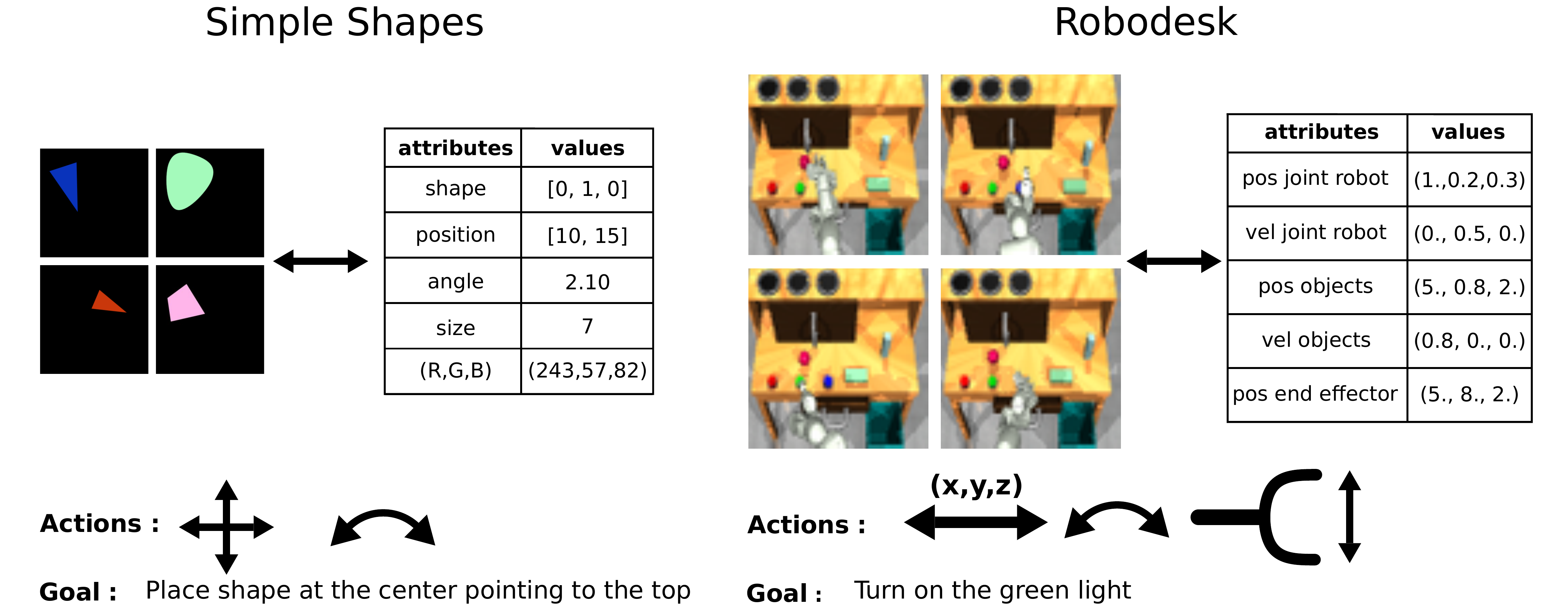}
    \end{center}
    \caption{Illustration of the environments and the tasks used in this study. For \textbf{Simple Shapes} (on the left), the Figure presents examples of raw observations, including four example images and one example set of attributes. The agent's goal is to place the shape at the center and pointing upward. The agent can move the shape one pixel at a time in four directions (up, down, left, right) or rotate it clockwise or counterclockwise by an angle of $\frac{\pi}{32}$. For \textbf{Robodesk} (on the right), the observations consist of fixed RGB images along with values representing the proprioception of the robotic arm and information about the objects in the scene. The actions are continuous and allow the robot arm to move, rotate, and open or close its end-effector. The agent's task is to turn on the green light by pressing the green button.}
    \label{Env}
\end{figure*}

The different models were tested in two different environments (Figure~\ref{Env}). The first one called `Simple Shapes', quite simple, was introduced in~\cite{devillers_semi-supervised_2024} as a fixed dataset, and extended in ~\cite{maytie_zero-shot_2024} as an RL environment. The second one called 'Robodesk', more complex, was introduced as a multi-task robotics benchmark by ~\cite{kannan2021robodesk}.

\subsection{Simple Shapes}

The Simple Shapes environment is multimodal, the agent can receive two types of observations: $32\times32$ pixel RGB images of a 2D shape on a black background, or a set of eight attributes directly describing the environment's state. There are three different types of shapes, an egg-like shape, an isosceles triangle, and a diamond. The shapes possess different properties: a size $s \in [s_{min}, s_{max}]$, a position $(x,y) \in [\frac{s_{max}}{2}, 32-\frac{s_{max}}{2}[^2$, a rotation $\theta \in [0, 2\pi[$ and an HSL color $(c_h,c_s,c_l) \in [0,1]^2 \times [l_{min},1]$. The agent does not observe these properties directly, but instead receives transformed attributes as observations: the rotation angle $\theta$ is decomposed into $(c_{\theta}, s_{\theta}) = (cos(\theta), sin(\theta))$; HSL colors are translated to the RGB domain, finally, the $shape$ variable is expressed as a one-hot vector of size three, and all variables are normalized between -1 and 1.

At the beginning of each episode, attributes are randomly sampled within their respective domains; the starting point is thus a random shape of a random orientation, located somewhere in the image. The agent's goal is to move the shape to the center of the image and align it to point to the top. For this purpose, six different actions are available to the agent: moving the shape by one pixel in cardinal directions (left, right, up, or down) and rotating the shape by an angle of $\frac{\pi}{32}$ clockwise or anti-clockwise. The reward is initialized at zero. At each timestep, the reward is equal to minus the current distance (in pixels) between the shape's position and the image center minus the smallest angle (in radians) between the shape's orientation and the null angle times a rotation reward coefficient equal to 10 by default. The episode ends when the shape reaches the goal state, with no additional reward.

Since this environment is relatively simple and each state can be reached with reasonable probability through random sampling, $500,000$ paired observations were collected randomly to train both the VAEs and the Global Workspace.

\subsection{RoboDesk}

Robodesk \cite{kannan2021robodesk} is a robotic environment built using the MuJoCo simulator, comprising a robotic arm and a desk with multiple objects placed on and around it. Two types of observations are provided. First, RGB images of size $64 \times 64$, captured from an almost top-down view that englobes both the desk and the robotic arm. Second, a vector (76 dimensions) representing the robot's proprioception, including joint positions, speeds, the end-effector's state (open or closed grip), as well as the positions and velocities of various objects in the scene.

At initialization, random noise is added to the default joint positions of the robot to get a plausible but randomized initial state. The objects in the scene are randomly positioned, except for the button and the bin. The drawer and sliding door are located at the same position but are more or less open.

This environment supports 9 distinct tasks as part of a multi-task benchmark, one of which was selected as the primary task for the present work (section~\ref{subsection:oneGW}; other tasks are considered as well in section~\ref{subsection:allGW}). In this task, the objective is for the robot to turn on the green light by pressing the green button. The reward function is a weighted sum that considers both whether the robot successfully pressed the green button and the distance between the end-effector and the button. To achieve this the robot has $2,000$ environment steps but with an action repetition of 4 steps, leading to $500$ chosen actions.

In this environment, many states are difficult to reach using a random policy. To address this, an expert agent was trained to solve the task and then used to collect non-random data. The expert agent was based on the Dreamer model of size L (i.e, 100M parameters), as outlined in \cite{hafner_mastering_2023}. A total of 1 million observations were randomly sampled from the environment, and another 1 million were sampled using the expert policy, resulting in a dataset of 2 million bimodal data samples that was used for pretraining the VAEs and Global Workspace.

\section{Results}

We evaluate our GW-Dreamer approach across multiple dimensions to demonstrate its effectiveness and practical advantages. First, we compare GW-Dreamer against various PPO and Dreamer baselines to establish its sample efficiency gains when combining Global Workspace representations with world model-based reinforcement learning. Second, we assess the robustness of our approach by examining how different models perform when one sensory modality is removed at test time, simulating real-world sensor failure conditions. Finally, we investigate the scalability and generalization capabilities of a single Global Workspace across multiple distinct tasks. Together, these experiments demonstrate that GW-Dreamer not only achieves superior sample efficiency but also provides the robustness and flexibility needed for practical deployment.

\subsection{One Global Workspace per task}
\label{subsection:oneGW}
We evaluated the performance of our GW-Dreamer model against different Proximal Policy Optimization (PPO) and Dreamer variants. GW, VAE and CLIP-like components were identical across all models and were pre-trained using data collected randomly or via an expert agent. All PPO algorithms were trained using the stable-baseline 3 implementation \cite{stable-baselines3}. Specifically, we trained: (1) a standard multimodal PPO model using raw observations as input, (2) a "VAE-PPO" that uses concatenated representations from both attributes and image VAEs, (3) a “CLIP-like-PPO” that uses a single multimodal representation as input, computed with the CLIP-like model, and (4) a "GW-PPO" that employs a single input representation coming from the GW. This last model is expected to have certain advantages relative to the previous three baselines, owing to its strong multimodal abilities (as shown by~\cite{maytie_zero-shot_2024}), but it does not include a World Model. PPO baselines are only shown for the Simple Shapes environment, because they did not succeed in solving the Robodesk task. Additionally, we trained (5) the standard Dreamer algorithm using both attribute and image modalities, and (6) a "VAE-Dreamer" model that receives VAE-based representations of attributes and images as inputs. The motivation for incorporating latent representations (VAEs) was to enable the models to operate in a fully latent space, reducing the high compute associated with reconstructing images through decoders. We also evaluated (7) a “CLIP-like-Dreamer” that receives a unique multimodal representation from a CLIP-like model; this was meant to compare our Global-Workspace model with another, more conventional multimodal representation system. Finally, we experimented with end-to-end training of our GW-Dreamer system: instead of pre-training the GW for the $\mathcal{L}_{cont}$ and $\mathcal{L}_{broad}$ objectives (Eq.~\ref{equation:gw_losses}) and then freezing its parameters during WM and AC policy training, we started from a randomly initialized GW and jointly trained its objectives together with WM and AC training.

\begin{figure*}[htb!]
    \begin{center}
        \includegraphics[scale=0.5]{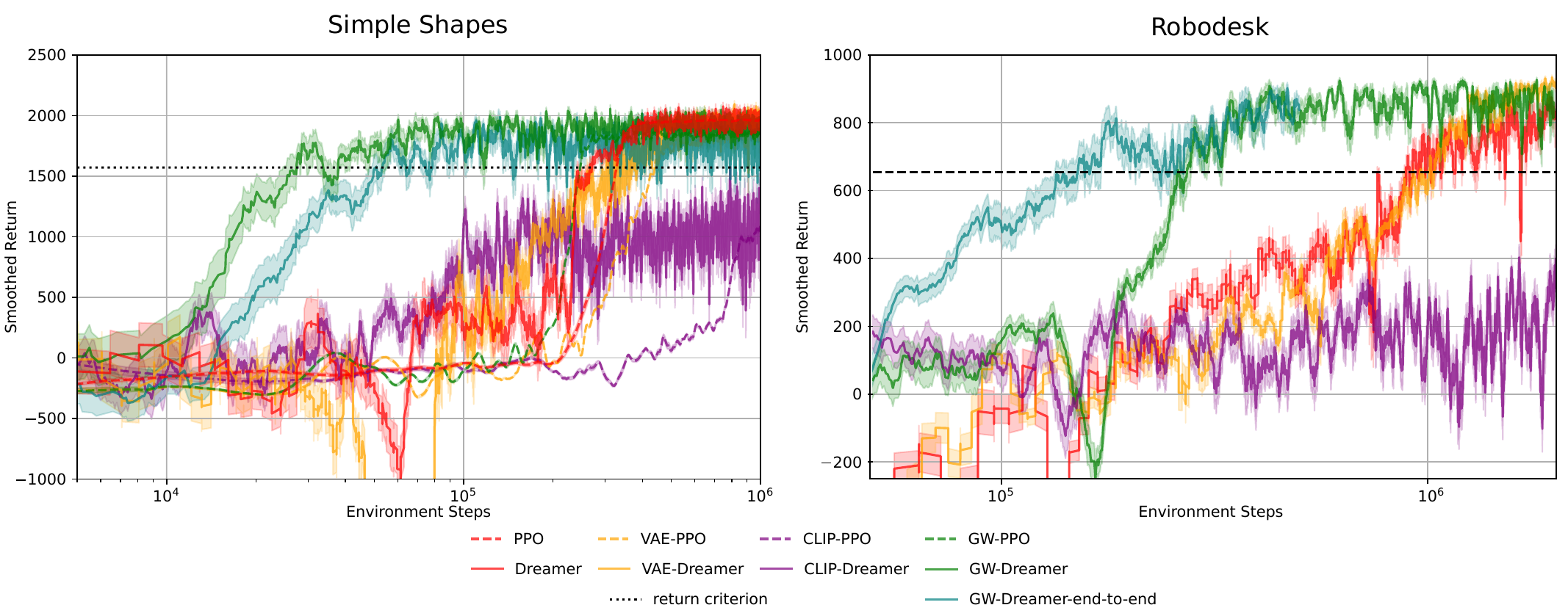}
    \end{center}
    \caption{Performance (cumulative sum of rewards or ``return'') as a function of the number of environment steps (log scale) during training in Simple Shapes environment on the left and Robodesk on the right. A fixed baseline, corresponding to the performance of a fully random policy, was subtracted from the episode returns. Thus, a random policy's performance is equal to zero. The returns are smoothed using a sliding window of length 10, with the shaded region indicating the standard error of the mean over this window. The return criterion is defined as 75\% of the maximum smoothed return in Simple Shapes and 70\% in Robodesk. It corresponds (as verified visually) to a performance at which the task starts to be solved properly.}
    \label{SS_res}
\end{figure*}

The results for both environments are illustrated in Figure \ref{SS_res}. Returns (cumulative sum of rewards) were normalized such that a return of zero corresponds to the performance of a random agent. Additionally, a “return criterion” was defined as respectively, 75\% (in Simple Shapes) and 70\% (in Robodesk) of the highest smoothed reward obtained by any of the models in the figure. We verified visually that this criterion corresponds to a level of return at which the task begins to be solved efficiently.

Figure \ref{SS_res} presents several key findings. First, among the PPO variants in Simple Shapes, GW-PPO reaches the return threshold the fastest, matching the convergence speed of the original Dreamer algorithm. Specifically, GW-PPO achieves the criterion at approximately 200,000 environment steps, while standard PPO and VAE-PPO reach it at around 300,000 and 400,000 steps, respectively. This speed gain is modest, highlighting the relatively limited interest of applying a GW representation on a world-model-free RL algorithm. Finally, CLIP-PPO converges significantly more slowly and attains a lower final return, suggesting that not all multimodal representation learning strategies are equally effective for reinforcement learning: GW training, with its semi-supervised broadcast objectives, appears superior to standard contrastive learning objectives~\cite{maytie_zero-shot_2024}.

Second, Dreamer-based models generally reach the return criterion of the Simple-Shapes task earlier than PPO-based ones, corroborating previous findings that world models improve sample efficiency \cite{hafner_mastering_2023}. Nonetheless, GW-PPO exhibits sample efficiency comparable to both standard Dreamer and VAE-Dreamer, with all three models reaching the threshold at approximately 200,000 environment steps. This suggests that a strong multimodal representation (GW) can bring as much to the model's efficiency as a World Model could. Can the two advantages (GW and World Model) be combined to yield an even more efficient architecture?

This is what Figure~\ref{SS_res} seems to suggest: the GW-Dreamer model outperforms all other models, reaching the criterion in just 20,000 steps in Simple Shapes, and 200,000 steps in Robodesk (i.e. about 10X and 4x faster than other Dreamer variants). Since both GW-Dreamer and VAE-Dreamer operate in a multimodal latent space, this result directly highlights the effectiveness of the GW latent representation in improving sample efficiency within a world-model framework. The figure also demonstrates the superiority of the GW representation over other multimodal strategies such as CLIP-like embeddings. While in Simple Shapes, CLIP-Dreamer converges significantly faster than CLIP-PPO, both models fail to reach the return threshold in both environments, revealing limitations of contrastive learning for multimodal integration.

Finally, we found that training the Global Workspace end-to-end alongside the rest of the model does not appear to hinder performance. In Simple Shapes, the GW-Dreamer-end-to-end variant reaches the threshold in approximately 50,000 steps, still 4 times faster than Dreamer or VAE-Dreamer, despite the added complexity of jointly optimizing multiple loss components during training. In RoboDesk, end-to-end training even appears to accelerate convergence, with the model reaching the performance threshold at 150,000 steps, compared to 200,000 steps for GW-Dreamer. This suggests that in more complex settings, where data coverage across all possible states is more challenging, allowing the Global Workspace to be trained jointly with the World Model and Actor-Critic Policy enables the model to construct a more effective multimodal representation. This, in turn, facilitates learning of the downstream components (WM, AC). Overall, these findings highlight that the GW representation enhances learning efficiency when integrated within a dreaming-based architecture.

While GW-Dreamer demonstrates superior sample efficiency, it is important to consider the computational overhead of its components. Table \ref{tab:FLOPs} presents the training FLOPs for different model components across all variants. The VAE components are shared between GW-Dreamer and VAE-Dreamer, while the Global Workspace architecture is used in both GW-Dreamer and CLIP-Dreamer, with the CLIP-like model employing only contrastive loss as the learning objective (no broadcast objective). The RL algorithm component (column ``RL 1 task'' in the table) represents the entire Dreamer algorithm for the baseline (listed as ``Dreamer 12M'' in the table), and only the World Model plus Actor-Critic components for GW-Dreamer, VAE-Dreamer, and CLIP-Dreamer.
Since RL training FLOPs are computed until each model reaches the return criterion, GW-Dreamer requires the fewest RL training FLOPs due to its dramatically reduced environment steps, $4 \times$ fewer in Robodesk compared to other Dreamer variants. However, when accounting for the pre-trained Global Workspace component, GW-Dreamer's total computational cost exceeds that of the original Dreamer. Importantly, GW pre-training represents a one-time computational cost that can be amortized across multiple downstream tasks; this could make the GW-Dreamer approach increasingly cost-effective as more tasks leverage the same representation. We explicitly test this prediction in section~\ref{subsection:allGW}.

\begin{table*}[!htb]
    \centering
    \resizebox{0.7\linewidth}{!}{
        \begin{tabular}{@{}lcccc||c@{}}
            \toprule
            {} & VAE & GW & RL 1 task & RL 6 tasks & RL 4 extra tasks\\
            \midrule
            GW-Dreamer   & $1.055 \times 10^{18}$ & $2.544\times 10^{16}$ & $5.410\times 10^{15}$ & $7.358 \times 10^{16}$ & $5.411 \times 10^{16}$\\
            Dreamer 12M  & N/A & N/A & $2.443\times 10^{16}$ & $2.044\times 10^{17}$ & $1.134 \times 10^{17}$\\
            Dreamer 40M  & N/A & N/A & X & $6.127\times 10^{18}$ & $4.158 \times 10^{18}$\\
            VAE-Dreamer  & $1.055 \times 10^{18}$ & N/A & $7.460 \times 10^{15}$ & X & X\\
            CLIP-Dreamer & $1.055 \times 10^{18}$ & $3.003\times 10^{15}$ & $9.516 \times 10^{15}$ & X & X\\
            \bottomrule
            \\
        \end{tabular}
     }
    \caption{Training FLOPs (forward and backward passes) for different model components in Robodesk environment. Rows represent different models, columns represent FLOPs for each component. VAE and GW FLOPs are computed until the checkpoint used for RL training. RL algorithm FLOPs are computed until the model reaches the return criterion defined in Figure~\ref{SS_res}. FLOPs for RL algorithm for the 6 tasks and 4 extra tasks are the sum of the FLOPs until the model convergence. "N/A" indicates that the model component (VAE or GW) is not applicable to that model, while "X" indicates that the model was not tested in that configuration. The count of FLOPs for the 4 extra tasks is separated from the rest since the Global Workspace and VAEs are applied zero-shot, as they have not been trained on expert data from these tasks; all numbers in that column are directly comparable.}
    \label{tab:FLOPs}
\end{table*}

\subsection{Modality Ablation Robustness}

One advantage of training a policy from two input modalities (images and attributes) is that the resulting agent could prove particularly robust in conditions where one of the two modalities becomes unreliable. We thus conducted an additional experiment to evaluate the zero-shot robustness of GW-Dreamer compared to other multimodal variants when one sensory modality is removed. This scenario simulates real-world conditions where a robot may experience sensor failure or a human may lose one of their senses. Once the models were trained, their parameters were frozen, and we systematically removed either the attribute or image inputs. For models using a GW (or CLIP-like) representation, the fusion mechanism was adjusted accordingly: if attributes were removed, full weight was assigned to the visual modality ($\alpha_v = 1, \alpha_{attr} = 0$), and conversely, if vision was removed the opposite adjustment was made ($\alpha_v = 0, \alpha_{attr} = 1$). An important thing to notice is that only the GW part of the model was trained to be robust to different fusion weights; during their training, the World Model and Actor-Critic always received GW (or CLIP-like) representations computed using equal fusion weights ($\alpha_v = \alpha_{attr} = 0.5$).

\begin{figure*}[htb!]
    \begin{center}
        \includegraphics[scale=0.6]{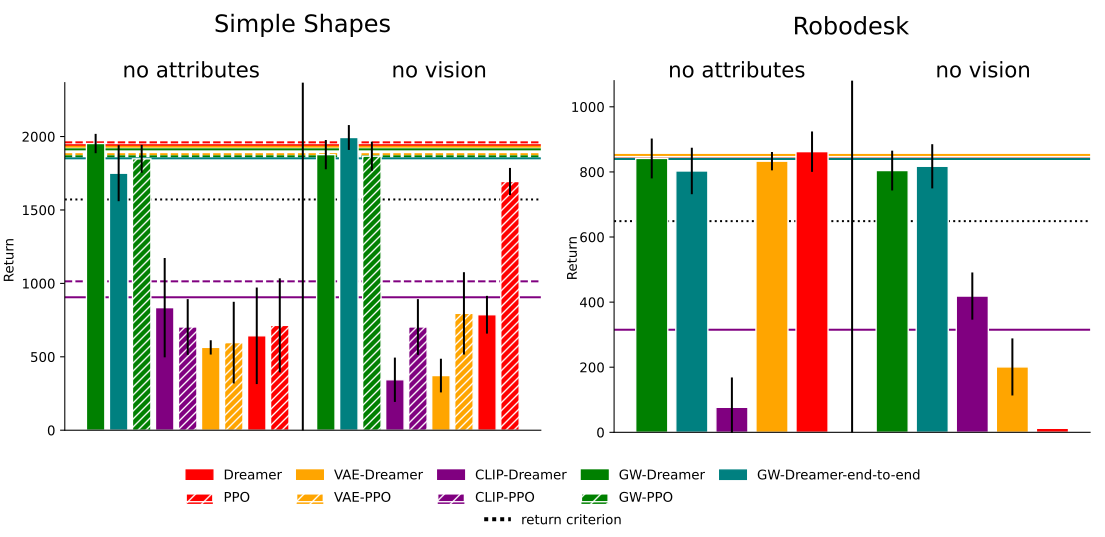}
    \end{center}
    \caption{Performance (return) of the different models under unimodal ablation for Simple Shapes on the left and Robodesk on the right. Solid bars represent Dreamer-based models, while hatched bars correspond to PPO-based models. In each plot, the left panel shows performance when the attribute modality is removed, and the right panel shows results following the removal of the vision modality. The return criterion, identical to the one used in Figure~\ref{SS_res}, represents the threshold above which the task is considered reliably solved. Colored horizontal lines indicate the maximum return achieved by each model in the full multimodal setting (from Figure~\ref{SS_res}), with solid lines for Dreamer variants and dashed lines for PPO variants. Bar and line colors are consistent across models for ease of comparison. Performance metrics represent the average over 10 independent runs, with error bars showing the standard error of the mean.}
    \label{SS_rob}
\end{figure*}

The results presented in Figure~\ref{SS_rob} reveal a striking contrast in model robustness under modality ablation. In both environments, GW-Dreamer, GW-Dreamer-end-to-end and GW-PPO maintain performance levels close to their original scores and remain above the return threshold, whereas all other models experience an important performance drop. Notably, in Simple Shapes, PPO without visual input shows a drop but still surpasses the return criterion, while it falls to chance without attributes. This suggests that, although trained with both visual and attribute modalities, standard PPO relies predominantly on attribute inputs, indicating it is not effectively leveraging multimodal information. In Robodesk, Dreamer and VAE-Dreamer display the same type of skewed behavior: while they were trained with both modalities, they suffer from the removal of vision but not attributes, indicating that they rely more on visual data than attributes.

Although CLIP-based models are trained to align modalities in a shared latent space, they exhibit limited robustness when one modality is removed at inference time (even though the fusion weights ($\alpha_v$,$\alpha_{attr}$) are adjusted to (1,0) or (0,1) in order to focus on the remaining modality). This is somewhat surprising, as we should expect aligned representations to allow the WM and AC components to function with either or both modalities present. In contrast, GW-based models demonstrate greater robustness under such conditions.

Overall, these results demonstrate that GW-Dreamer is capable of solving the task even when one modality is missing, underscoring the importance of multimodal representations like GW that support adaptive integration of sensory inputs. While GW-PPO also benefits from this representation, it remains less sample-efficient than GW-Dreamer.

\subsection{Unique Global Workspace for all tasks}
\label{subsection:allGW}

While we showed that the GW-Dreamer RL algorithm converged faster than Dreamer and other algorithms, the total number of training FLOPs was not advantageous when considering a single task, because of the overhead cost of GW pretraining. However, once a Global Workspace multimodal representation has been pretrained, we can potentially leverage it to encode observations across multiple tasks rather than only one. This could make the GW-Dreamer approach computationally more efficient overall.

To demonstrate this, we collected expert data for 6 different tasks composing the Robodesk environment, which resulted in a dataset of 3.5 million samples (500,000 per task and 500,000 using a random policy) containing images and attribute vectors. For convenience, these data were collected using a pre-trained Dreamer-V3 model with 100 million parameters, but could alternatively have been collected through human demonstrations. (Note that 3 fine-grained manipulation tasks out of all 9 tasks in the Robodesk environment could not be learned, even using a large 100M-parameter Dreamer-V3 model; so we only consider here the remaining 6 tasks, for which we could collect expert data). Using the method explained in Section~\ref{sec:Model}, we trained the vision and attributes VAEs and the Global Workspace model using these data. We then used this single pre-trained Global Workspace across all six tasks. We trained six different GW-Dreamer instances sharing a single Global Workspace model, thereby amortizing the computational cost of pre-training the Global Workspace. For comparison, we also trained six different Dreamers, one for each task, each time using a Dreamer composed of either 12 million parameters (the number of trainable parameters in GW-Dreamer) or 40 million parameters (the total number of parameters in GW-Dreamer).

In these experiments, we report success count rather than cumulative reward, as the reward functions are not always reliable indicators of goal achievement in Robodesk tasks (e.g., for tasks requiring placing an object in a bin or off the table, part of the reward is based on the distance between the end effector and the object, a metric that becomes irrelevant once the task is successfully completed). The success count represents the number of steps within an episode during which the task criterion is satisfied. Specifically, at each step, the metric returns 1 if the task goal is achieved and 0 otherwise, and these values are summed over the entire episode. Since all episodes have a fixed length of 500 steps (without early termination upon task completion), the maximum possible success count is 500.

\begin{figure*}[!htbp]
    \begin{center}
        \includegraphics[width=\textwidth]{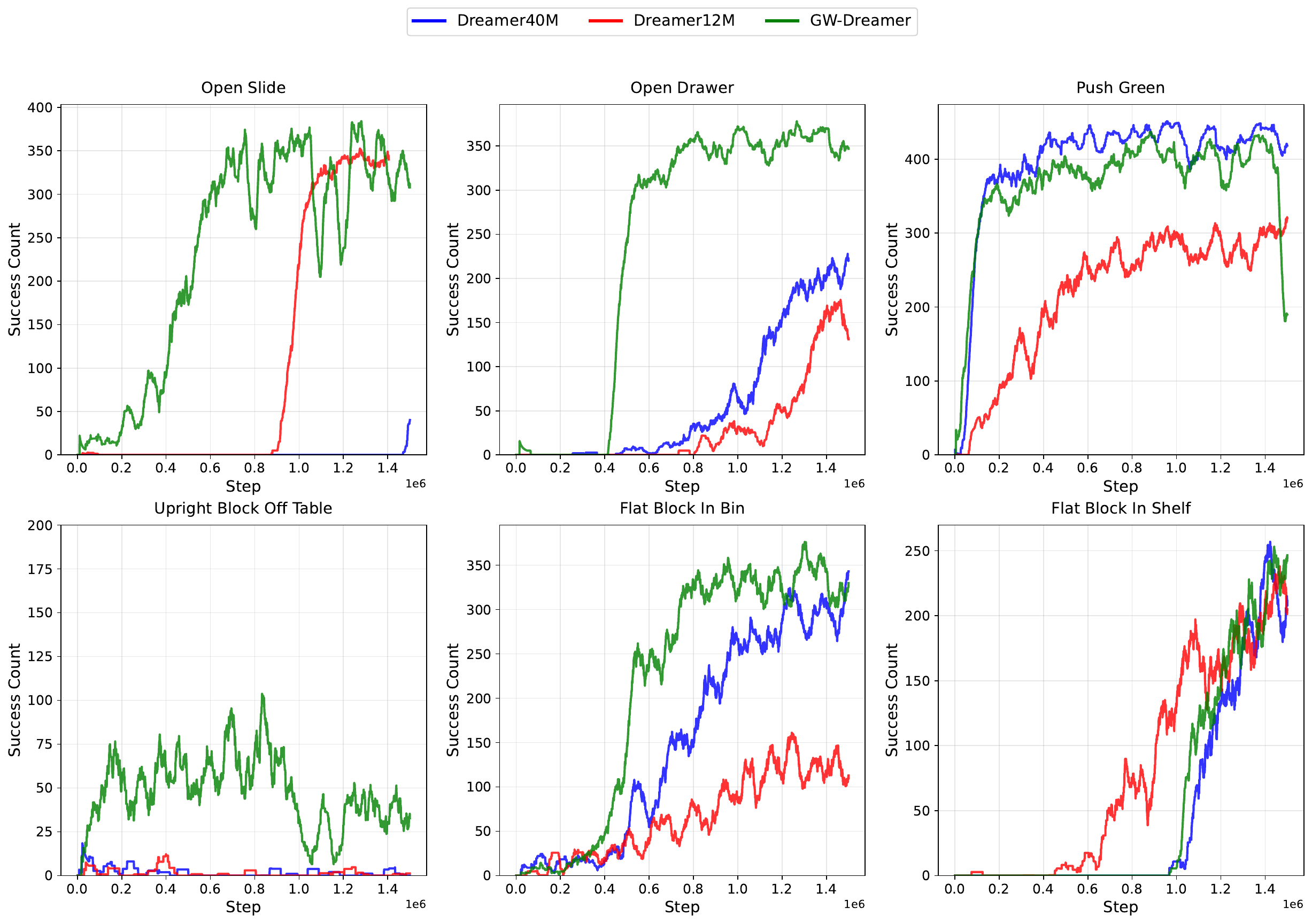}
    \end{center}
    \caption[Performance comparison across 6 Robodesk tasks for GW-Dreamer]{Performance comparison across 6 Robodesk tasks. The y-axis represents the success count (number of steps per episode in which the task was successfully completed). The number of steps per episode is common to all the tasks and fixed to 500 steps. The x-axis shows the number of environment steps during training (in millions). Results are smoothed using a sliding window of length 100.}
    \label{GWDreamer6tasks}
\end{figure*}

Figure~\ref{GWDreamer6tasks} presents the results obtained for each task. GW-Dreamer outperforms or converges faster than Dreamer 12M in 5 of the 6 tasks. As shown in Table~\ref{tab:FLOPs} (column ``RL 6 tasks''), the number of FLOPs required during RL training is significantly lower for GW-Dreamer, although the pre-training phase still represents a substantial computational cost. Overall, including GW pretraining costs, the GW-Dreamer model requires half the training resources of the Dreamer 12M model, while achieving higher performance on the 6 tasks. When comparing our model with Dreamer 40M, the performance gap narrows considerably, particularly for the "push green" and "flat block in bin" tasks. However, this performance improvement comes at a considerable cost in training FLOPs, as illustrated in Table~\ref{tab:FLOPs}. Despite this, GW-Dreamer still performs better than or equal to Dreamer 40M while requiring 60 times fewer total training FLOPs when considering GW training costs, and 6 times fewer when also accounting for the pre-training cost of the VAEs. This demonstrates the advantage of using a pre-trained Global Workspace that has learned to represent a multimodal environment, enabling the model to scale effectively without substantially increasing training FLOPs while maintaining strong performance.

\subsection{Extra tasks}
 
The shared training FLOPs required to train a single Global Workspace model on six tasks amortize the pre-training costs for each task. We also tested the capacity of GW-Dreamer, relying on a pre-trained Global Workspace model, to solve tasks unseen by the Global Workspace during training. For this experiment, we used the same pre-trained Global Workspace trained on 3.5 million data samples from the six previous tasks and applied it zero-shot to new, unseen extra tasks (these tasks were provided as part of the Robodesk environment, but not included in the multi-task benchmark). The rest of the model (World Model and Actor-Critic) was trained as explained in Section~\ref{sec:Model} by interacting directly with the environment, as in the original Dreamer algorithm. We compared GW-Dreamer against Dreamer models with 12M and 40M parameters.

\begin{figure*}[!htbp]
    \begin{center}
        \includegraphics[width=\textwidth]{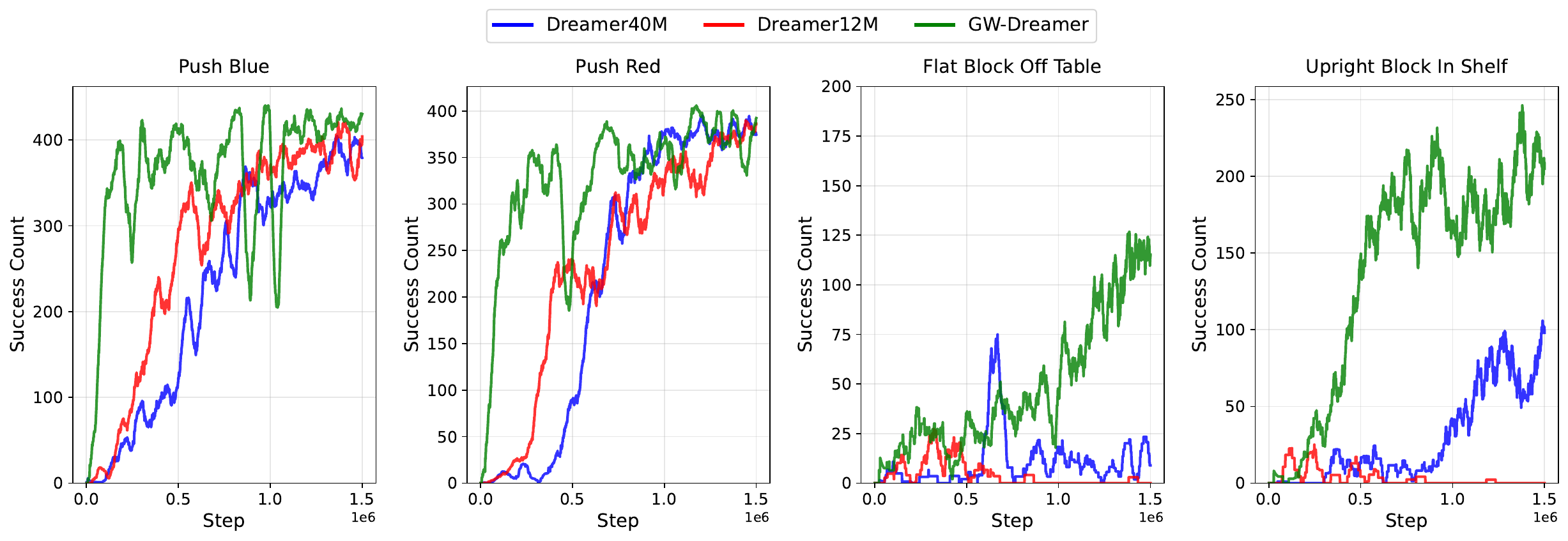}
    \end{center}
    \caption[Performance comparison across 4 Robodesk extra tasks for GW-Dreamer]{Performance comparison across 4 Robodesk extra tasks. The same Global Workspace and VAEs as those used in Figure~\ref{GWDreamer6tasks} were applied to these new tasks zero-shot, without training or fine-tuning on expert data from these tasks. Only the World-Model and Actor-Critic were trained end-to-end by interacting with the environment. The y-axis represents the success count (number of steps per episode in which the task was successfully completed). The number of steps per episode is common to all the tasks and fixed to 500 steps. The x-axis shows the number of environment steps during training (in millions). Results are smoothed using a sliding window of length 100.}
    \label{GWDreameroodtasks}
\end{figure*}

Figure~\ref{GWDreameroodtasks} presents the results obtained for four extra tasks (Robodeks provides 9 extra tasks overall, but five tasks involving objects manipulation could not be solved by any of the models we considered, and are therefore not included here). In this setup, even though its Global Workspace never observed expert demonstrations of these tasks, GW-Dreamer is systematically superior to both Dreamer baselines: it converges in fewer environment steps for two tasks, and reaches higher success count on the two other tasks.  

Using a pre-trained Global Workspace that has not seen expert demonstrations for the extra tasks during training illustrates how the Global Workspace could serve as a "foundation" model capable of encoding environment observations into compact latent representations. This setup also allows us to focus the FLOP comparison on RL algorithms (World Model and Actor-Critic) exclusively, as the Global Workspace and VAE components were applied zero-shot to these unseen tasks. Table~\ref{tab:FLOPs} (column ``RL 4 extra tasks'') illustrates that in this case, the number of FLOPs required during training for GW-Dreamer is significantly (2 to 75 times) lower than for the 12M and 40M parameter DreamerV3 models.

\section{Discussion and Conclusion}
This paper represents a first step in bridging Global Workspace Theory and World Models in AI. It builds upon the architecture proposed by \cite{vanrullen_deep_2021} and implemented by \cite{devillers_semi-supervised_2024}, adapting it to be compatible with World-Model-based reinforcement learning algorithms such as Dreamer~\cite{ha_recurrent_2018, hafner_mastering_2023}.

Despite its contributions, this study has several limitations. First, it relies on pre-trained Global Workspace (GW) models and pre-trained VAEs, which were trained either on randomly sampled data or on data collected by an expert agent. The last case introduces an additional constraint: access to either pre-existing environment data or an expert policy capable of generating such data. A possible solution to this issue could be the use of pretrained foundation models that can encode arbitrary visual inputs into latent spaces (e.g., \cite{oquab_dinov2_2023}), or the development of custom foundation encoders trained on large-scale, publicly available robotics datasets \cite{walke_bridgedata_2023, collaboration_open_2024}. The current work attempts to mitigate this limitation by also introducing an end-to-end training strategy, where only the VAEs are trained in advance on pre-collected data, while the GW is trained jointly with the rest of the system. This design maintains consistency with the GW framework proposed by \cite{vanrullen_deep_2021}, which advocates for using pretrained unimodal encoders as GW modules, while offering greater flexibility, such as the possibility of collecting different modalities independently, or substituting them with general-purpose encoders.
As a case in point, the recently released Dreamer 4 model \cite{hafner_training_2025} makes use of pre-trained models on large-scale datasets. Our work aligns with this idea of pre-training "foundation" models for an environment, as demonstrated by the capacity of GW-Dreamer to achieve strong performance while relying on a pre-trained Global Workspace trained on only a subset of possible tasks.

Another limitation concerns the complexity of the evaluation environments. While DreamerV3 has demonstrated remarkable performance on highly complex tasks like Minecraft \cite{hafner_mastering_2023}, the Simple Shapes and Robodesk environments tested in this study are comparatively simpler. Nevertheless, this range allows for a preliminary assessment of the scaling properties of the proposed architecture, from a minimal setup to a more challenging, higher-dimensional multi-task benchmark.

Despite these limitations, this study highlights the benefits of integrating GWT with WM. This integration substantially enhances RL training efficiency, enabling GW-Dreamer to achieve comparable performance using approximately 4 to 10 times fewer environment interactions than DreamerV3--a state-of-the-art RL system already known for its remarkable training efficiency~\cite{hafner_mastering_2023}. Notably, this acceleration is not solely due to operating in a pretrained latent space, as GW-Dreamer also outperforms a Dreamer variant that utilizes VAE-based latent representations as inputs. The results further underscore the advantages of GW over other multimodal strategies, such as contrastive learning-based methods.
The pre-training of a separate Global Workspace for each task is not advantageous in terms of training FLOPs, but the use of a single Global Workspace across several downstream RL tasks demonstrates the possibility of improving RL training efficiency in GW-Dreamer. This is further supported by the capacity of GW-Dreamer to achieve strong performance and train faster than classical Dreamer, even on tasks unseen by the Global Workspace. In these scenarios, GW-Dreamer maintains more efficient training and sometimes outperforms Dreamer in terms of task success, while requiring fewer training FLOPs. This highlights the capacity of the Global Workspace model to serve as a "foundation" model for encoding high-dimensional observations into latent representations usable by a World Model or an RL policy.
In addition, GW-Dreamer exhibits strong zero-shot robustness to modality loss. Unlike the original Dreamer, GW-Dreamer maintains stable performance even when either visual inputs or attribute information are removed, a property also observed in model-free RL settings \cite{maytie_zero-shot_2024}, and corroborated in this work using the GW-PPO variant. By contrast, a multimodal representation model trained using contrastive objectives (e.g., CLIP-like) under the same conditions does not demonstrate similar robustness to missing modalities.
Currently, this robustness is facilitated by manually adapting fusion weights at inference time. A promising direction for future work would be to enable the model to autonomously select which modalities to route into the Global Workspace, based on which modality is missing. Another interesting extension would be to test the robustness not only to missing modalities but to various types and intensities of noise.

The findings also have potential implications in cognitive neuroscience as a practical test of GWT, a prominent theory of multimodal integration and high-level cognition. First, compared with existing approaches~\cite{radford_learning_2021,hafner_mastering_2023}, the GW tends to produce a superior multimodal representation, owing to its semi-supervised training procedure inspired by the ``broadcast'' principle~\cite{baars_cognitive_1988}. This fact was already suggested by a number of recent studies~\cite{devillers_does_2021, devillers_semi-supervised_2024,maytie_zero-shot_2024}, and is confirmed here in the context of model-based RL. Second, we show that GW multimodal representations can be leveraged by a World Model to produce mental simulations that help the system converge to an optimal decision strategy. This resembles ``dreaming'' in humans and animals, and more generally, captures the ability of such biological systems to imagine the potential outcome of a planned sequence of actions before making a decision. While this evidently does not suffice to validate GWT, it confirms its potential relevance as a theory of higher-level cognition.

Ultimately, this research tackles key challenges in RL, such as the large amount of environment interactions required for policy training and the need for strong multimodal representations, particularly in robotics. It opens the way for future work to further integrate GWT and World Models.

\section{Acknowledgments}
This work was supported by an ANITI Chair (ANR grant ANR-19-PI3A-004), an ANR grant COCOBOT (ANR21-FAI2-0005) and by ``Défi Clé Robotique centrée sur l'humain'' funded by Région Occitanie, France. This research is also funded by the European Union (ERC Advanced GLOW project number 101096017). Views and opinions expressed are however those of the authors only and do not necessarily reflect those of the European Union or the European Research Council Executive Agency. Neither the European Union nor the granting authority can be held responsible for them.

\bibliography{CCN2025}
\bibliographystyle{IEEEtran}

\newpage
\clearpage

\appendices

\section{VAE details}
\subsection{Architecture}

This section details the architecture chosen for the VAEs used in the different models (Global-Workspace, VAE-Dreamer, VAE-PPO). They are encoding raw observations ($o^v, o^{attr}$) to latent unimodal representation ($z^v, z^{attr}$). The visual VAEs, detailed in Tables \ref{table:VAE_v_SS} and \ref{table:VAE_v_Rob}, are composed of Convolutional Layers with Batch Norm and ReLU. The latent dimension is of size 10 for Simple Shapes and 40 for Robodesk. The total number of parameters is about $5,8M$ for Simple Shapes and $11,2M$ for Robodesk. The attributes VAEs, detailed in Tables \ref{table:VAE_attr_SS} and \ref{table:VAE_attr_Rob}, are much smaller, with 11,000 parameters for Simple Shapes and $1,1M$ for Robodesk. It is composed of Multiple Linear and ReLU layers with a latent dimension of 10 for Simple Shapes and 40 for Robodesk. In Simple Shapes, the last layer of the decoder is divided in two parts, one of size 3 to predict the class of the shape (one-hot encoding), another one of size 8 with a Tanh activation to predict the other attributes.

\begin{table}[htb!]
    \begin{center}
    \resizebox{\columnwidth}{!}{
        \begin{tabular}{l|l}
    VAE encoder ($2.8M$ params) & VAE decoder ($3M$ params)\\
    \hline
    $x\in \mathbb{R}^{3\times 32\times 32}$ & $z\in \mathbb{R}^{10}$ \\
    $\text{Conv}_{128} - \text{BN} - \text{ReLU}$ & $\text{FC}_{8\times8\times 1024}$ \\
    $\text{Conv}_{256} - \text{BN} - \text{ReLU}$ & $\text{ConvT}_{512}-\text{BN}-\text{ReLU}$ \\
    $\text{Conv}_{512} - \text{BN} - \text{ReLU}$ & $\text{ConvT}_{256}-\text{BN}-\text{ReLU}$ \\
    $\text{Conv}_{1024} - \text{BN} - \text{ReLU}$ & $\text{ConvT}_{128}-\text{BN}-\text{ReLU}$ \\
    $\text{Flatten} - \text{FC}_{2\times 10}$ & $\text{Conv}_{1}-\text{Sigmoid}$ \\
        \end{tabular}
    }
    \end{center}
    \caption{Architecture and number of parameters of the visual VAE for 'Simple Shapes'.}
    \label{table:VAE_v_SS}
\end{table}

\begin{table}[htb!]
    \begin{center}
    \resizebox{\columnwidth}{!}{
        \begin{tabular}{l|l}
    VAE encoder ($11.5M$ params) & VAE decoder ($13.6M$ params)\\
    \hline
    $x\in \mathbb{R}^{3\times 64\times 64}$ & $z\in \mathbb{R}^{40}$ \\
    $\text{Conv}_{64} - \text{BN} - \text{ReLU}$ & $\text{FC}_{8\times8\times1024}$ \\
    $\text{Conv}_{128} - \text{BN} - \text{ReLU}$ & $\text{ConvT}_{1024}-\text{BN}-\text{ReLU}$ \\
    $\text{Conv}_{256} - \text{BN} - \text{ReLU}$ & $\text{ConvT}_{256}-\text{BN}-\text{ReLU}$ \\
    $\text{Conv}_{512} - \text{BN} - \text{ReLU}$ & $\text{ConvT}_{128}-\text{BN}-\text{ReLU}$ \\
    $\text{Conv}_{1024} - \text{BN} - \text{ReLU}$ & $\text{ConvT}_{128}-\text{BN}-\text{ReLU}$ \\
    $\text{Flatten} - \text{FC}_{2\times 40}$ & $\text{Conv}_{1}-\text{Sigmoid}$ \\
        \end{tabular}
    }
    \end{center}
    \caption{Architecture and number of parameters of the visual VAE for 'Robodesk'.}
    \label{table:VAE_v_Rob}
\end{table}

\begin{table}[htb!]
    \begin{center}
    \resizebox{\columnwidth}{!}{
        \begin{tabular}{l|l}
    VAE encoder ($6,700$ params) & VAE decoder ($4,700$ params)\\
    \hline
    $x\in \mathbb{R}^{11}$ & $z\in \mathbb{R}^{10}$ \\
    $\text{FC}_{64} - \text{ReLU}$ & $\text{FC}_{64} - \text{ReLU}$ \\
    $\text{FC}_{64} - \text{ReLU}$ & $\text{FC}_{64} - \text{ReLU}$ \\
    $\text{FC}_{10} - \text{ReLU}$ & $\text{FC}_{3} \times \text{FC}_{8} - \text{Tanh}$ \\
    $\text{FC}_{2\times 10}$       &  \\
        \end{tabular}
    }
    \end{center}
    \caption{Architecture and number of parameters of the attributes VAE for 'Simple Shapes'.}
    \label{table:VAE_attr_SS}
\end{table}

\begin{table}[htb!]
    \begin{center}
    \resizebox{\columnwidth}{!}{
        \begin{tabular}{l|l}
    VAE encoder ($618,000$ params) & VAE decoder ($851,000$ params)\\
    \hline
    $x\in \mathbb{R}^{76}$ & $z\in \mathbb{R}^{40}$ \\
    $\text{FC}_{512} - \text{ReLU}$ & $\text{FC}_{512} - \text{ReLU}$ \\
    $\text{FC}_{512} - \text{ReLU}$ & $\text{FC}_{512} - \text{ReLU}$ \\
    $\text{FC}_{512} - \text{ReLU}$ & $\text{FC}_{512} - \text{ReLU}$ \\
    $\text{FC}_{512} - \text{ReLU}$ & $\text{FC}_{512} - \text{ReLU}$ \\
    $\text{FC}_{40} - \text{ReLU}$ & $\text{FC}_{[40,36]} $ \\
        \end{tabular}
    }
    \end{center}
    \caption{Architecture and number of parameters of the attributes VAE for 'Robodesk'.}
    \label{table:VAE_attr_Rob}
\end{table}

\subsection{Training}

The training objective for the VAEs is identical across both modalities and environments. It corresponds to the standard $\beta$-VAE loss function:

$\mathcal{L}_{\beta\text{-VAE}} = \mathbb{E}_{q_\phi(\mathbf{z}|\mathbf{x})}[\log p_\theta(\mathbf{x}|\mathbf{z})] - \beta \cdot D_{\mathrm{KL}}\left(q_\phi(\mathbf{z}|\mathbf{x}) \,\|\, p(\mathbf{z})\right)$\\

where $\mathbf{x}$ denotes the input data, $\mathbf{z}$ is the latent variable, $q_\phi(\mathbf{z}|\mathbf{x})$ represents the encoder (i.e., the approximate posterior), $p_\theta(\mathbf{x}|\mathbf{z})$ is the decoder (i.e., the likelihood), and $p(\mathbf{z})$ is the prior distribution, typically chosen as a standard Gaussian. The term $D_{\mathrm{KL}}$ denotes the Kullback–Leibler divergence, and the hyperparameter $\beta \geq 0$ modulates the weight of the KL regularization term. The objective is to learn a compact latent representation that follow a Gaussian prior while enabling accurate reconstruction of the input data. Training details are provided in Table~\ref{table:VAE_training}.

\begin{table}[htb!]
    \begin{center}
    \resizebox{\columnwidth}{!}{
        \begin{tabular}{l|l|l}
    & Simple Shapes & Robodesk\\
    \hline
    Number of data & $500.000$ & $3.000.000$\\
    Batch size & $2.056$ & $2.056$ \\
    Number of training steps & $500.000$ & $500.000$\\
    Optimizer & AdamW & AdamW \\
    LR Scheduler & OneCycleLR & OneCycleLR \\
    max learning rate & visual: $5e^{-3}$ & visual: $3e^{-4}$\\
                      & attributes: $5e^{-3}$ & attributes: $5e^{-4}$ \\
    start / end learning rate & $5e^{-3}$ & $1e^{-4}$ \\
    weight decay & $1e^{-5}$ & $1e^{-5}$ \\
        \end{tabular}
    }
    \end{center}
    \caption{Hyperparameters used during VAEs training for both environments}
    \label{table:VAE_training}
\end{table}

\section{GW details}
\subsection{Architecture}

This section details the architecture used in the Global Workspace. Tables \ref{table:GW_SS} and \ref{table:GW_Rob} gives details about encoders ($e_v$, $e_{attr}$) and decoders ($d_v$, $d_{attr}$) architecture of the Global Workspace for both environments. Because they are identical for both modalities only one table provides the architecture’s details. The encoders and decoders are simply a sequence of Linear layers with ReLU activation function, and the latent dimension of the GW is of size 10 for Simple Shapes and 60 for Robodesk. This GW model takes inputs from the VAEs described before and is used in the GW-Dreamer and GW-PPO models.

\begin{table}[htb!]
    \begin{center}
        \resizebox{\columnwidth}{!}{
            \begin{tabular}{l|l}
        GW encoder ($13,800$ params) & GW decoder ($13,800$ params)\\
        \hline
        $x\in \mathbb{R}^{10}$ & $z\in \mathbb{R}^{10}$ \\
        $\text{FC}_{64} - \text{ReLU}$ & $\text{FC}_{64} - \text{ReLU}$ \\
        $\text{FC}_{64} - \text{ReLU}$ & $\text{FC}_{64} - \text{ReLU}$ \\
        $\text{FC}_{64} - \text{ReLU}$ & $\text{FC}_{64} - \text{ReLU}$ \\
        $\text{FC}_{64} - \text{ReLU}$ & $\text{FC}_{64} - \text{ReLU}$ \\
        $\text{FC}_{10} $ & $\text{FC}_{10}$ \\
            \end{tabular}
        }
    \end{center}
    \caption{Architecture and number of parameters of the encoders and decoders of the Global Workspace model for one modality in 'Simple Shapes'.}
    \label{table:GW_SS}
\end{table}

\begin{table}[htb!]
    \begin{center}
        \resizebox{\columnwidth}{!}{
            \begin{tabular}{l|l}
        GW encoder ($585,000$ params) & GW decoder ($585,000$ params)\\
        \hline
        $x\in \mathbb{R}^{40}$ & $z\in \mathbb{R}^{76}$ \\
        $\text{FC}_{512} - \text{ReLU}$ & $\text{FC}_{512} - \text{ReLU}$ \\
        $\text{FC}_{512} - \text{ReLU}$ & $\text{FC}_{512} - \text{ReLU}$ \\
        $\text{FC}_{512} - \text{ReLU}$ & $\text{FC}_{512} - \text{ReLU}$ \\
        $\text{FC}_{512} - \text{ReLU}$ & $\text{FC}_{512} - \text{ReLU}$ \\
        $\text{FC}_{76} $ & $\text{FC}_{40}$ \\
            \end{tabular}
        }
    \end{center}
    \caption{Architecture and number of parameters of the encoders and decoders of the Global Workspace model for one modality in 'Robodesk'.}
    \label{table:GW_Rob}
\end{table}

\subsection{Training}

The losses used to train the model are described in Equations \ref{L_broadcast} and \ref{L_GW}. These losses are scaled by different weights as shown in Table \ref{table:GW_losses}. These values were taken from the implementation done by \cite{devillers_semi-supervised_2024}, where the weight of the contrastive loss (measured by a cross-entropy function) was always smaller compared to the other ones (measured using MSE distance). Other training details are provided in Table~\ref{table:GW_training}. The training data used to train the Global Workspace are the same than the one used to train the VAE described above.

\begin{table}[htb!]
    \begin{center}
        \begin{tabular}{l|l|l}
    & GW & CLIP-like\\
    \hline
    $\beta{dcy}$ & 1 & 0\\
    $\beta{cy}$ & 1 & 0\\
    $\beta{tr}$ & 1 & 0\\
    $\beta{fusion}$ & 1 & 0\\
    $\beta{cont}$ & 0.1 & 1\\
        \end{tabular}
    \end{center}
    \caption{GW losses weights used to train a Global Workspace model and a CLIP-like model}
    \label{table:GW_losses}
\end{table}

\begin{table}[htb!]
    \begin{center}
    \resizebox{\columnwidth}{!}{
        \begin{tabular}{l|l|l}
    & Simple Shapes & Robodesk\\
    \hline
    Number of data & $500.000$ & $3.000.000$\\
    Batch size & $2.056$ & $2.056$ \\
    Number of training steps & $500.000$ & $500.000$\\
    Optimizer & AdamW & AdamW \\
    LR Scheduler & OneCycleLR & OneCycleLR \\
    max learning rate & $5e^{-3}$ & $3e^{-4}$\\
    start / end learning rate & $5e^{-4}$ & $1e^{-4}$ \\
    weight decay & $1e^{-5}$ & $1e^{-6}$ \\
        \end{tabular}
    }
    \end{center}
    \caption{Hyperparameters used during GW training for both environments}
    \label{table:GW_training}
\end{table}

\section{GW-Dreamer details}
\subsection{Architecture}
This section provides details on the architectures and parameters used for the GW-Dreamer model in both environments. The model consists of two main components: a World Model and an Actor-Critic module.

The World Model includes a dynamic component implemented as a single-layer GRU, which receives as input a concatenation of the Global Workspace (GW) representation and the action vector, details are provided in Table \ref{table:GRU_archi}. It is followed by three distinct output heads, each designed to extract specific information from the GRU’s latent state. Architectural details of these heads are provided in Table~\ref{table:WM_heads_SS} for the Simple Shapes environment and Table~\ref{table:WM_heads_Rob} for Robodesk. The head responsible for predicting the next GW latent vector is a simple linear layer followed by a Tanh activation function. The other two heads—responsible for predicting the scalar reward and a Boolean termination flag (done)—share the same architecture, consisting of a sequence of linear layers interleaved with ReLU activation functions.

The Actor-Critic component follows the design proposed in Dreamer~\cite{hafner_mastering_2023}. Both the Actor and Critic networks are composed of linear layers with ReLU activations. The Actor outputs the parameters (mean and variance) of a probability distribution over possible actions, while the Critic estimates a distribution over the value of the current state. As in the original Dreamer, the output distributions are discretized using a fixed number of bins, allowing the use of a categorical cross-entropy loss between one-hot encoded targets and the predicted softmax distributions. Detailed architecture and parameter values are listed in Table~\ref{table:AC_SS} for Simple Shapes and Table~\ref{table:AC_Rob} for Robodesk.

An overview of the parameters repartition composing the model are visible in Table \ref{table:sizes}. The 

\begin{table}[htb!]
    \begin{center}
        \begin{tabular}{l|l|l}
    & Simple Shapes & Robodesk\\
    \hline
    GRU parameters & $26,000$ & $6.3M$ \\
    GRU input size & $32$ & $1024$ \\
    GRU hidden size & $64$ & $1024$ \\
        \end{tabular}
    \end{center}
    \caption{Details about the GRU architecture in both environments}
    \label{table:GRU_archi}
\end{table}

\begin{table}[htb!]
    \begin{center}
    \resizebox{\columnwidth}{!}{
        \begin{tabular}{l|l|l}
    GW head ($650$ params) & reward head ($150,000$ params) & done head ($85,000$ params)\\
    \hline
    $x\in \mathbb{R}^{64}$ & $z\in \mathbb{R}^{74}$ & $z\in \mathbb{R}^{74}$\\
    $\text{FC}_{10} - \text{Tanh}$ & $\text{FC}_{256} - \text{ReLU}$ & $\text{FC}_{256} - \text{ReLU}$ \\
                                   & $\text{FC}_{256} - \text{ReLU}$ & $\text{FC}_{256} - \text{ReLU}$ \\
                                   & $\text{FC}_{256} - \text{ReLU}$ & $\text{FC}_{256} - \text{ReLU}$ \\
                                   & $\text{FC}_{256}$ & $\text{FC}_{1}$ \\
        \end{tabular}
    }
    \end{center}
    \caption{Architecture and number of parameters of the different heads composing the World Model for the Simple Shapes environment.}
    \label{table:WM_heads_SS}
\end{table}

\begin{table}[htb!]
    \begin{center}
    \resizebox{\columnwidth}{!}{
        \begin{tabular}{l|l|l}
    GW head ($826,000$ params) & reward head ($413,000$ params) & done head ($348,000$ params)\\
    \hline
    $x\in \mathbb{R}^{1024}$ & $z\in \mathbb{R}^{1100}$ & $z\in \mathbb{R}^{1100}$ \\
    $\text{FC}_{512} - \text{ReLU}$ & $\text{FC}_{256} - \text{ReLU}$  & $\text{FC}_{256} - \text{ReLU}$\\
    $\text{FC}_{512} - \text{ReLU}$ & $\text{FC}_{256} - \text{ReLU}$ & $\text{FC}_{256} - \text{ReLU}$\\
    $\text{FC}_{76} - \text{Tanh}$ & $\text{FC}_{256} - \text{ReLU}$ & $\text{FC}_{256} - \text{ReLU}$\\
                                   & $\text{FC}_{256}$ & $\text{FC}_{1}$ \\
        \end{tabular}
    }
    \end{center}
    \caption{Architecture and number of parameters of the different heads composing the World Model for the Robodesk environment.}
    \label{table:WM_heads_Rob}
\end{table}

\begin{table}[htb!]
    \begin{center}
        \begin{tabular}{l|l}
        Actor ($500 K$ params) & Critic ($800 K$ params)\\
    \hline
    $x\in \mathbb{R}^{64}$ & $z\in \mathbb{R}^{64}$ \\
    $\text{FC}_{512} - \text{ReLU}$ & $\text{FC}_{512} - \text{ReLU}$ \\
    $\text{FC}_{512} - \text{ReLU}$ & $\text{FC}_{512} - \text{ReLU}$ \\
    $\text{FC}_{512} - \text{ReLU}$ & $\text{FC}_{512} - \text{ReLU}$ \\
    $\text{FC}_{2\times 7}$         & $\text{FC}_{2\times 255}$ \\
        \end{tabular}
    \end{center}
    \caption{Architecture and number of parameters of the Actor-Critic for Simple Shapes.}
    \label{table:AC_SS}
\end{table}

\begin{table}[htb!]
    \begin{center}
        \begin{tabular}{l|l}
        Actor ($1.1M$ params) & Critic ($1.2M$ params)\\
    \hline
    $x\in \mathbb{R}^{512}$ & $z\in \mathbb{R}^{512}$ \\
    $\text{FC}_{512} - \text{ReLU}$ & $\text{FC}_{512} - \text{ReLU}$ \\
    $\text{FC}_{512} - \text{ReLU}$ & $\text{FC}_{512} - \text{ReLU}$ \\
    $\text{FC}_{512} - \text{ReLU}$ & $\text{FC}_{512} - \text{ReLU}$ \\
    $\text{FC}_{512} - \text{ReLU}$ & $\text{FC}_{512} - \text{ReLU}$ \\
    $\text{FC}_{2\times 5}$         & $\text{FC}_{2\times 255}$ \\
        \end{tabular}
    \end{center}
    \caption{Architecture and number of parameters of the Actor-Critic for Robodesk.}
    \label{table:AC_Rob}
\end{table}

\begin{table}[htb!]
    \begin{center}
        \begin{tabular}{l|l|l}
    & Simple Shapes & Robodesk\\
    \hline
    WM & $313K$ & $7.8M$\\
    AC & $1,9M$ & $3.7M$\\
    VAE (frozen) & $5.7M$ & $26.5M$\\
    GW (frozen) & $55K$ & $2.3M$\\
    \hline
    Total size & $7.35M$ & $39M$\\
    \hline
    Trainable size & $2.2M$ & $11.4M$ \\
        \end{tabular}
    \end{center}
    \caption{Size of the different network that are trainable or not composing the GW-Dreamer model}
    \label{table:sizes}
\end{table}

\subsection{Training}

Table \ref{table:GW_Dreamer_training} summarizes the key hyperparameters used to train GW-Dreamer. The overall training procedure is detailed in Algorithm \ref{algo:training_algorithm}. When training the Global Workspace, World Model, and Actor-Critic jointly (GW-Dreamer-end-to-end), all hyperparameters are kept the same. The Global Workspace is trained as described previously, with optimizer settings specified in Table \ref{table:GW_Dreamer_GW_training}. It is updated concurrently with the other components (every n environment steps), but for each global training step, the Global Workspace performs two training updates.

\begin{table}[htb!]
    \begin{center}
        \begin{tabular}{l|l|l}
    & Simple Shapes & Robodesk\\
    \hline
    Buffer Size & $1e^5$ & $1e^5$\\
    Batch Size & $64$ & $32$ \\
    Batch Length & $32$ & $128$ \\
    Train Ratio & $4$ & $2$ \\
    Imagination length & $20$ & $20$ \\
    \hline
    Optimizer & Adam & Adam \\
    LR & WM: $1e^{-3}$ & WM: $3e^{-4}$ \\
    & AC: $3e^{-4}$ & AC: $3e^{-4}$ \\
    Return lambda & $0.95$ & $0.95$ \\
        \end{tabular}
    \end{center}
    \caption{Hyperparameters used during GW-Dreamer training for both environments}
    \label{table:GW_Dreamer_training}
\end{table}

\begin{table}[htb!]
    \begin{center}
        \begin{tabular}{l|l|l}
    & Simple Shapes & Robodesk\\
    \hline
    Optimizer & AdamW & AdamW\\
    LR Scheduler & OneCycleLR & OneCycleLR \\
    start LR & $3e^{-3}$ & $1e^{-4}$ \\
    max LR & $1e^{-2}$ & $3e^{-4}$ \\
    end LR & $1e^{-3}$ & $1e^{-4}$ \\
    cycle size & $25,000$ & $50,000$
        \end{tabular}
    \end{center}
    \caption{Hyperparameters used for the training of the Global Workspace in GW-Dreamer End-to-End for both environments}
    \label{table:GW_Dreamer_GW_training}
\end{table}

\section{Dreamer details}
\subsection{Architecture}

The architecture of the Dreamer algorithm is the same as the one introduced in \cite{hafner_mastering_2023}. The size of the architecture is adapted to match the GW-Dreamer setting in terms of number of trainable parameters. The details are given in Table \ref{table:Dreamer_sized}.

\begin{table}[htb!]
    \begin{center}
        \begin{tabular}{l|l|l|l}
    & Simple Shapes & Robodesk 12M & Robodesk 40M \\
    \hline
    GRU recurrent units & $512$ & $2048$ & $4096$ \\
    Dense hidden units & $128$ & $256$ & $512$\\
    RSSM hidden dimension & $128$ & $384$ & $384$\\
    Total size & $2M$ & $12M$ & $40M$\\
        \end{tabular}
    \end{center}
    \caption{Architecture details used in Dreamer baseline}
    \label{table:Dreamer_sized}
\end{table}

\subsection{Training}

The training hyperparameters are also kept as similar as possible to the original ones from \cite{hafner_mastering_2023}. The changes are made to have similar learning rates and a close number of training data at each training step. The details are given in Table \ref{table:Dreamer_training}. In the case of Robodesk putting a learning rate of $1e^{-3}$ leads to more instability and more steps to reach the criterion, while keeping it at $3e^{-4}$ was improving the results.

\begin{table}[htb!]
    \begin{center}
        \begin{tabular}{l|l|l}
    & Simple Shapes & Robodesk\\
    \hline
    Buffer Size & $1e^6$ & $1e^6$\\
    Batch Size & $16$ & $16$ \\
    Batch Length & $65$ & $65$ \\
    Train Ratio & $32$ & $32$ \\
    Imagination length & $15$ & $15$ \\
    \hline
    Optimizer & Adam & Adam \\
    LR & WM: $1e^{-3}$ & WM: $3e^{-4}$\\
    & AC: $3e^{-4}$ & AC: $3e^{-4}$\\
    Return lambda & $0.95$ & $0.95$ \\
        \end{tabular}
    \end{center}
    \caption{Hyperparameters used during Dreamer training for both environments}
    \label{table:Dreamer_training}
\end{table}

\section{Algorithms}

The algorithm used to train GW-Dreamer is presented in Algorithm \ref{algo:training_algorithm}. In this case the unimodal models (VAEs) and the GW are pre-trained while the WM and AC are trained together.

\begin{algorithm}[H]
    \centering
    \caption{GW-Dreamer Training Procedure}
    \begin{algorithmic}[1]
        \STATE \textbf{Require} Pretrained Global Workspace (GW)
        \STATE \textbf{Initialize} World Model (WM), Actor-Critic (AC), Replay Buffer
        \WHILE{not max number environment steps}
            \STATE Collect $n$ transitions $(o_t, a_t, o_{t+1}, r_{t+1}, d_{t+1})$ from the environment and store in Replay Buffer
            \FOR{$m$ training steps}
                \STATE \textbf{Train World Model:}
                \STATE \quad Transform Replay Buffer observations $(o_t^v, o_t^{attr})$ into GW latent vectors $(z_t)$
                \STATE \quad Predict next GW representation, reward and done signal: $WM(z_t,a_t) = (\hat{d}_{t+1}, \hat{r}_{t+1}, \hat{z}_{t+1})$
                \STATE \quad Compute $\mathcal{L}_{WM}$ by comparing against $(d_{t+1}, r_{t+1}, z_{t+1})$
                \STATE \quad Update WM by backpropagating $\mathcal{L}_{WM}$
                \STATE \textbf{Train Actor-Critic:}
                \STATE \quad Encode the first observation through GW to get latent representation $z_t$
                \STATE \quad Generate imagined trajectories using WM and the actions predicted by AC
                \STATE \quad Update AC using simulated rewards and transitions
            \ENDFOR
        \ENDWHILE
    \end{algorithmic}
    \label{algo:training_algorithm}
\end{algorithm}

The algorithm used to train GW-Dreamer End-to-End is presented in Algorithm \ref{algo:training_algorithm_e2e}. In this case only the unimodal models (VAEs) are pre-trained and the Global Workspace is initialized randomly.

\begin{algorithm}[ht!]
    \centering
    \caption{GW-Dreamer End-to-End Training Procedure}
    \begin{algorithmic}[1]
        \STATE \textbf{Require} Pretrained unimodal models (VAEs)
        \STATE \textbf{Initialize} Global Workspace (GW), World Model (WM), Actor-Critic (AC), Replay Buffer
        \WHILE{not max number environment steps}
            \STATE Collect $n$ transitions $(o_t, a_t, o_{t+1}, r_{t+1}, d_{t+1})$ from the environment and store in Replay Buffer
            \FOR{$m$ training steps}
                \FOR{$i$ training steps}
                    \STATE \textbf{Train Global Workspace:}
                    \STATE \quad Encode the different observations using the pre-trained VAEs to unimodal latent representations
                    \STATE \quad Encode the different unimodal latents inside the GW (z) with random and fixed fusion weights depending on the loss ($\alpha_{attr}, \alpha_{v}$).
                    \STATE \quad Decode the GW (z) back to the different unimodal latent representations
                    \STATE \quad Compute the different losses (contrastive, demi-cycles, cycles, translation, fusion) and sum them
                \ENDFOR
                \STATE \textbf{Train World Model:}
                \STATE \quad Transform Replay Buffer observations $(o_t^v, o_t^{attr})$ into GW latent vectors using the update GW $(z_t)$
                \STATE \quad Predict next GW representation, reward and done signal: $WM(z_t,a_t) = (\hat{d}_{t+1}, \hat{r}_{t+1}, \hat{z}_{t+1})$
                \STATE \quad Compute $\mathcal{L}_{WM}$ by comparing against $(d_{t+1}, r_{t+1}, z_{t+1})$
                \STATE \quad Update WM by backpropagating $\mathcal{L}_{WM}$
                \STATE \textbf{Train Actor-Critic:}
                \STATE \quad Encode the first observation through GW to get latent representation $z_t$
                \STATE \quad Generate imagined trajectories using WM and the actions predicted by AC
                \STATE \quad Update AC using simulated rewards and transitions
            \ENDFOR
        \ENDWHILE
    \end{algorithmic}
    \label{algo:training_algorithm_e2e}
\end{algorithm}

\section{Implementation Details}

GW-Dreamer is implemented in Pytorch 2.1 following the code introduced by \cite{devillers_semi-supervised_2024} and available at \footnote{\url{https://github.com/ruflab/shimmer}}. Dreamer was tested using the code implemented in JAX by \cite{hafner_mastering_2023} in \footnote{\url{https://github.com/danijar/dreamerv3}}. The training of the smallest models (PPO, GW-Dreamer, CLIP-Dreamer in Simple Shapes) were performed on a Nvidia Quadro RTX 5000 16GB. All the other models were trained on a Nvidia A100 80GB.

\end{document}